# Compositional Model Repositories via Dynamic Constraint Satisfaction with Order-of-Magnitude Preferences

**Jeroen Keppens** JEROEN@INF.ED.AC.UK
**Qiang Shen** QIANGS@INF.ED.AC.UK
*School of Informatics, The University of Edinburgh*
*Appleton Tower, Crichton Street, Edinburgh EH8 9LE, UK*

## Abstract

The predominant knowledge-based approach to automated model construction, compositional modelling, employs a set of models of particular functional components. Its inference mechanism takes a scenario describing the constituent interacting components of a system and translates it into a useful mathematical model. This paper presents a novel compositional modelling approach aimed at building model repositories. It furthers the field in two respects. Firstly, it expands the application domain of compositional modelling to systems that can not be easily described in terms of interacting functional components, such as ecological systems. Secondly, it enables the incorporation of user preferences into the model selection process. These features are achieved by casting the compositional modelling problem as an activity-based dynamic preference constraint satisfaction problem, where the dynamic constraints describe the restrictions imposed over the composition of partial models and the preferences correspond to those of the user of the automated modeller. In addition, the preference levels are represented through the use of symbolic values that differ in orders of magnitude.

## 1. Introduction

Mathematical models form an important aid in understanding complex systems. They also help problem solvers to capture and reason about the essential features and dynamics of such systems. Constructing mathematical models is not an easy task, however, and many disciplines have contributed approaches to automate it. Compositional modelling (Falkenhainer & Forbus, 1991; Keppens & Shen, 2001b) is an important class of approaches to automated model construction. It uses predominantly knowledge-based techniques to translate a high level scenario into a mathematical model. The knowledge base usually consists of generic fragments of models that provide one of the possible mathematical representation of a process that occurs in one or more components. The inference mechanisms instantiate this knowledge base, search for the most appropriate selection of model fragments, and compose them into a mathematical model. Compositional modelling has been successfully applied to a variety of application domains ranging from simple physics, over various engineering problems to biological systems.

The present work aims at a compositional modelling approach for building model repositories of ecological systems. In the ecological modelling literature, a range of models have been devised to formally characterise the various phenomena that occur in ecological systems. For example, the logistic growth (Verhulst, 1838) and the Holling predation (Holling, 1959) models describe the changes in the size of a population. The former expresses changes due to births and deaths and the latter changes due to one population feeding on another. A compositional model repository aims to make such (partial) models more generally usable by providing a mechanism to instantiate and compose them into larger models for more complex systems involving many interacting phenomena.

Thus, the input to a compositional model repository is a scenario describing the configuration of a system to be modelled. A sample scenario may include a number of populations and various predation and competition relations between them. The output is a mathematical model, called a scenario model, representing the behaviour of the system specified in the given scenario. A set of differential equations describing the changes in the population sizes in the aforementioned scenario due to births, natural deaths, deaths because of predation, available food supply or competition would constitute such a scenario model.

This application domain poses three important new challenges to compositional modelling. Firstly, the processes and components of an ecological system that are to be represented in the resulting composed model depend on one another and on the ways they are described. In population dynamics for example, models describing the predation or competition phenomena between two populations rely on the existence of a population growth model for each of the populations involved in the phenomenon. This inhibits the conventional approach of searching for a consistent and adequate combination of partial models, one for each component in the scenario. This approach provides an adequate solution for physical systems because these are comprised of components implementing a particular functionality that can be described by one or multiple partial models. Although the seminal work on compositional modelling (Falkenhainer & Forbus, 1991) recognised the existence of more complex interdependencies in model construction in general, it provided only a partial solution for it: all the conditions under which certain modelling choices were relevant had to be specified manually in the knowledge base.

Secondly, the domain of ecology lacks a complete theory of what constitutes an adequate model. Most existing compositional modellers are based on a predefined concept of model adequacy. They employ inference mechanisms that are guaranteed to find a model that meets such adequacy criteria. However, criteria to determine how adequate an ecological model may be vary between ecological domains and even between the ecologists that require the model within the same domain. Therefore, the compositional modeller requires a facility to define the properties that the generated ecological models must satisfy.

Thirdly, it is not possible to express all the criteria imposed on the scenario model in terms of hard requirements. Often, ecological models that describe mechanisms and behaviours are only partially understood. In such cases, the choice of one model over another becomes a matter of expert opinion rather than pure theory. Therefore, in the ecological domain, modelling approaches and presumptions are, to some extent, selected based on preferences. Existing compositional modellers are not equipped to deal with such user preferences and this paper presents the very first compositional modeller that incorporates them.

Generally speaking, the above three issues are tackled in this paper by means of a method to translate the compositional modelling problem into an activity-based dynamic preference constraint satisfaction problem (aDPCSP) (Keppens & Shen, 2002). An aDPCSP integrates the concept of activity-based dynamic constraint satisfaction problem (aDCSP) (Miguel & Shen, 1999; Mittal & Falkenhainer, 1990) with that of order-of-magnitude preferences (Keppens & Shen, 2002). The attributes and domains of this aDPCSP correspond to model design decisions, with constraints describing the restrictions imposed by consistency requirements and properties and order-of-magnitude preferences describing the user's preferences on modelling choices. The translation method brings the additional advantage that compositional modelling problems can now be solved by means of efficient aDCSP techniques. As such, compositional modellers can benefit from recent and future advances in constraint satisfaction research.

The remainder of this paper is organised as follows. Section 2 introduces the concept of an aDPCSP, a preference calculus that is suitable to express subjective user preferences for model design decisions and to be integrated with the general framework of aDPCSPs. It also gives a solution algorithm for aDPCSPs. Next, section 3 presents the compositional model repository and shows how such an aDPCSP is employed for automated model construction. These theoretical ideas are then illustrated by means of a large example in section 4, applying the compositional model repository to population dynamics problems. Section 5 concludes this paper with a summary and an outline of further research.

## 2. Dynamic Constraint Satisfaction with Order-of-Magnitude Preferences

In this section, a preference calculus based on order-of-magnitude reasoning is introduced and integrated into the activity-based dynamic constraint satisfaction problem (aDCSP) to form an aDCSP with order-of-magnitude preferences (aDPCSP). Then, a solution algorithm for such aDPCSPs is presented. The theory is illustrated with examples from the compositional modelling domain.

### 2.1 Background: Activity-based dynamic preference constraint satisfaction

A *hard constraint satisfaction problem* (CSP) is a tuple $\langle \mathbf{X}, \mathbf{D}, \mathbf{C} \rangle$, where

- $\mathbf{X} = \{x_1, \ldots, x_n\}$ is a vector of n attributes,
- $\mathbf{D} = \{D_{x_1}, \ldots, D_{x_n}\}$ is a vector containing exactly one domain for each attribute in $\mathbf{X}$. Each domain $D_x \in \mathbf{D}$ is a set of values $\{d_{i1}, \ldots, d_{in_i}\}$ that may be assigned to the attribute corresponding to the domain.
- $\mathbf{C}$ is a set of compatibility constraints. A compatibility constraint $c_{\{x_i,\ldots,x_j\}} \in \mathbf{C}$ defines a relation over a subset of the domains $D_{x_i}, ..., D_{x_j}$, and hence $c_{\{x_i,\ldots,x_j\}} \subseteq D_{x_i} \times \ldots \times D_{x_j}$.

A *solution* to a hard constraint satisfaction problem is any tuple $\langle x_1 : d_{x_1}, \ldots, x_n : d_{x_n} \rangle$ such that

- each attribute is assigned a value from its domain: $\forall x_i \in \mathbf{X}, d_{x_i} \in D_{x_i}$, and
- all compatibility constraints are satisfied: $\forall x_{\{x_i,\ldots,x_j\}} \in \mathbf{C}, \langle d_{x_i}, \ldots, d_{x_j} \rangle \in c_{\{x_i,\ldots,x_j\}}$.

An activity-based dynamic CSP (aDCSP), originally proposed in by Mittal and Falkenhainer (1990), extends conventional CSPs with the notion of activity of attributes. In an aDCSP, not all attributes are necessarily assigned in a solution, but only the active ones. As such, each attribute is either active and assigned a value or inactive:

$$\forall x_i \in \mathbf{X}, \left(\exists d_{x_i} \in D_{x_i}, x_i : d_{x_i}\right) \leftrightarrow \text{active}(x_i)$$

The activity of attributes in an aDCSP is governed by *activity constraints* that enforce under which assignments of attributes, an assignment to another attribute is relevant or possible. This information is important because it not only dictates for which attributes a value must be searched, but also the set of compatibility constraints that must be satisfied. Clearly, only the compatibility constraints $c_{\{x_i,\ldots,x_j\}} \in \mathbf{C}$ for which all attributes $x_i, \ldots, x_j$ are active must be satisfied, and a hard CSP is a sub-type of aDCSP in which all attributes are always active.

In summary, an *activity-based dynamic constraint satisfaction problem* (aDCSP) is a tuple $\langle \mathbf{X}, \mathbf{D}, \mathbf{C}, \mathbf{A} \rangle$, where

- $\langle \mathbf{X}, \mathbf{D}, \mathbf{C} \rangle$ is a hard CSP, and
- $\mathbf{A}$ is a set of activity constraints. An activity constraint restricts the sets of attribute-value assignments under which an attribute is active or inactive:

$$a_{x_i,\{x_j,\ldots,x_k\}} \subseteq D_{x_j} \times \ldots \times D_{x_k} \times \{\text{active}(x_i), \neg\text{active}(x_i)\}$$

  where $x_i \notin \{x_j, \ldots, x_k\}$.

A *solution* to an activity-based dynamic constraint satisfaction problem is any tuple $\langle x_1 : d_{x_1}, \ldots, x_l : d_{x_l} \rangle$ such that

- the attributes that are part of the solution are assigned a value from their domain: $\forall x_i \in \{x_1, \ldots, x_l\}, d_{x_i} \in D_{x_i}$,
- all activity constraints are satisfied:

$$\begin{aligned}
&\forall a_{x_i,\{x_j,\ldots,x_k\}} \in \mathbf{A}, \big(x_j \notin \{x_1, \ldots, x_l\}\big) \vee \ldots \vee \big(x_k \notin \{x_1, \ldots, x_l\}\big) \vee \\
&\big(x_i \in \{x_1, \ldots, x_l\} \wedge \langle d_{x_j}, \ldots, d_{x_k}, \text{active}(x_i) \rangle \in a_{x_i,\{x_j,\ldots,x_k\}}\big) \vee \\
&\big(x_i \notin \{x_1, \ldots, x_l\} \wedge \langle d_{x_j}, \ldots, d_{x_k}, \neg\text{active}(x_i) \rangle \in a_{x_i,\{x_j,\ldots,x_k\}}\big)
\end{aligned}$$

  and
- all compatibility constraints are satisfied:

$$\forall c_{\{x_i,\ldots,x_j\}} \in \mathbf{C}, \neg\text{active}(x_i) \vee \ldots \vee \neg\text{active}(x_j) \vee \langle d_{x_i}, \ldots, d_{x_j} \rangle \in c_{\{x_i,\ldots,x_j\}}$$

### 2.2 Order-of-magnitude preferences (OMPs)

Although an aDCSP can capture the hard constraints over decisions in a given problem as well as their dynamically changing solution space (as described by the activity constraints), the representation scheme it employs does not take into account any preferences users may have over possible alternative value assignments. Therefore, this work is extended to allow preference information to be attached to attribute-value assignments. The way in which this can be achieved depends on the representation and reasoning mechanisms underlying the preference calculus. In general, a preference calculus can be defined as a tuple $\langle \mathbb{P}, \oplus, \preccurlyeq \rangle$ where:

- $\mathbb{P}$ is the set of preferences,
- $\oplus$ is a commutative, associative operator that is closed in $\mathbb{P}$, and
- $\preccurlyeq$ forms a partial order, that is, reflexive, anti-symmetric and transitive relation defined over $\mathbb{P} \times \mathbb{P}$.

Because $\preccurlyeq$ is reflexive, antisymmetric and transitive, comparing preferences with the $\preccurlyeq$ relation yields one of four possible results:

- Two preferences $P_1, P_2 \in \mathbb{P}$ are equal to one another (denoted $P_1 = P_2$) iff $P_1 \preccurlyeq P_2$ and $P_2 \preccurlyeq P_1$.
- A preference $P_1 \in \mathbb{P}$ is strictly greater than a preference $P_2 \in \mathbb{P}$ (denoted $P_1 \succ P_2$) iff $P_1 \not\preccurlyeq P_2$ and $P_2 \preccurlyeq P_1$.
- A preference $P_1 \in \mathbb{P}$ is strictly smaller than a preference $P_2 \in \mathbb{P}$ (denoted $P_1 \prec P_2$) iff $P_1 \preccurlyeq P_2$ and $P_2 \not\preccurlyeq P_1$.
- Two preferences $P_1, P_2 \in \mathbb{P}$ are incomparable with one another (denoted $P_1 ? P_2$) iff $P_1 \not\preccurlyeq P_2$ and $P_2 \not\preccurlyeq P_1$.

Thus, an *activity-based dynamic preference constraint satisfaction problem* (aDPCSP) is a tuple $\langle \mathbf{X}, \mathbf{D}, \mathbf{C}, \mathbf{A}, \langle \mathbb{P}, \oplus, \preccurlyeq \rangle, P \rangle$ where

- $\langle \mathbf{X}, \mathbf{D}, \mathbf{C}, \mathbf{A} \rangle$ is an aDCSP,
- $\langle \mathbb{P}, \oplus, \preccurlyeq \rangle$ is a preference calculus, and
- $P$ is a mapping $D_{x_1} \cup \ldots \cup D_{x_n} \mapsto \mathbb{P}$ from the individual attribute-value assignments to the preferences.

The preferences attached to attribute-value assignments express the relative desirability of these assignments. The aim of the aDPCSP is to find a solution with the highest combined preference. That is, given an aDPCSP $\langle \mathbf{X}, \mathbf{D}, \mathbf{C}, \mathbf{A}, \langle \mathbb{P}, \oplus, \preccurlyeq \rangle, P \rangle$, any solution $\langle x_i : d_{x_i}, \ldots, x_j : d_{x_j} \rangle$ of the aDCSP $\langle \mathbf{X}, \mathbf{D}, \mathbf{C}, \mathbf{A} \rangle$ such that no other solution $\langle x_k : d_{x_k}, \ldots, x_l : d_{x_l} \rangle$ of $\langle \mathbf{X}, \mathbf{D}, \mathbf{C}, \mathbf{A} \rangle$ exists with $P(x_i : d_{x_i}) \oplus \ldots \oplus P(x_j : d_{x_j}) \prec P(x_k : d_{x_k}) \oplus \ldots \oplus P(x_l : d_{x_l})$ is a *solution* to the aDPCSP.

In this section, a preference calculus is introduced to extend an aDCSP into an aDPCSP. The calculus will be illustrated with examples from the compositional modelling domain.

### 2.2.1 REPRESENTATION OF OMPS

Technically, OMPs are combinations of so-called *basic preference quantities* (BPQs), which are the primitive units of preference or utility valuation associated with possible design decisions. Because it is often difficult to evaluate these BPQs numerically, they are ordered relative to one another employing similar ordering relations as those employed by relative order-of-magnitude calculi (Dague, 1993a, 1993b).

Let $\mathbb{B}$ be the set of all BPQs with respect to a particular decision problem. The BPQs in $\mathbb{B}$ are ordered with respect to one another at two levels of granularity, by two relations $\ll$ and $<$. First, $\mathbb{B}$ is partitioned into orders of magnitude, which are ordered by $\ll$. Then, the BPQs within each order of magnitude are ordered by $<$. Formally, an *order-of-magnitude ordering* over BPQs $\mathbb{B}$ is a tuple $\langle \mathbf{O}, \ll \rangle$, where $\mathbf{O} = \{O_1, \ldots, O_q\}$ is a partition of $\mathbb{B}$ and $\ll$ is an irreflexive and transitive binary relation over $\mathbf{O}$. Any subset of BPQs $O \in \mathbf{O}$ is said to be an *order of magnitude* in $\mathbb{B}$. Similarly, a *within-magnitude ordering* over a set of BPQs is a tuple $\langle O, < \rangle$, where $O$ is an order of magnitude in $\mathbb{B}$ and $<$ is an irreflexive and transitive binary relation over $O$.

To illustrate these ideas, consider the problem of constructing an ecological model describing a scenario containing a number of populations. Let some of the populations be parasites and others be hosts for these parasites. Also, assume that certain populations compete with others for scarce resources. In order to construct a scenario model, the compositional modeller must make a number of model design decisions: which population growth, host-parasitoid and competition phenomena are relevant, and which types of model best describe these phenomena.

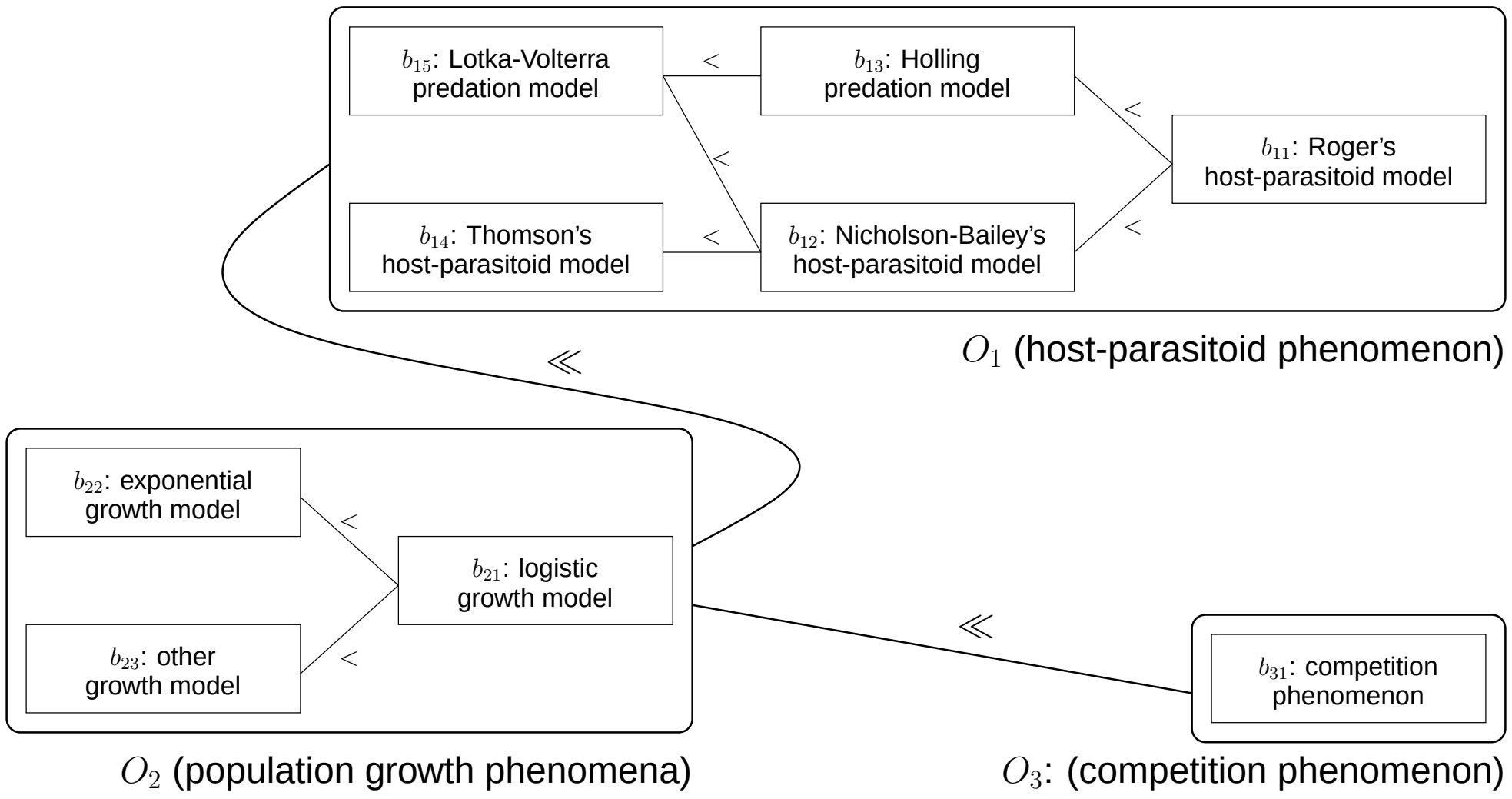

Figure 1: Sample space of BPQs $\mathbb{B}$

Figure 1 shows a sample space of BPQs that correspond to the selection of types of model. For the sake of illustration, the presumption is made that the quality of a scenario model depends on the inclusion of types of model, rather than on the inclusion or exclusion of phenomena. Apart from $b_{23}$ and $b_{31}$, all BPQs correspond to standard textbook ecological models[1]. BPQ $b_{23}$ stands for the use of a population growth model that is implicit in another population growth model (the Lotka-Volterra model, for instance, implicitly includes its own concept of growth). Finally, BPQ $b_{31}$ is the preference associated with a competition model (say, the only one included in the knowledge base).

The 9 BPQs in this sample space are partitioned over 3 orders of magnitude. The $\ll$ relation orders the orders of magnitude: $O_2 \ll O_1$ and $O_2 \ll O_3$. The binary $<$ relation orders individual BPQs within an order of magnitude. In the BPQ ordering within $O_1$, for instance, Rogers' host-parasitoid model ($b_{11}$) is preferred over that by Nicholson and Bailey ($b_{12}$) and the Holling predation model ($b_{13}$). The latter two models can not be compared with one another, but they both are preferred over the Lotka-Volterra model. Furthermore, Thompson's host-parasitoid model is less preferred than that of Nicholson and Bailey, but it can not be compared with the Lotka-Volterra and Holling models.

#### 2.2.2 COMBINATIONS OF OMPS

By definition, OMPs are combinations of BPQs. The implicit value of an OMP $p$ equals the combination $b_1 \oplus \ldots \oplus b_n$ of its constituent BPQs $b_1, \ldots, b_n$. This property allows OMPs to be defined as functions such that an OMP $P = b_1 \oplus \ldots \oplus b_n$ is a function $f_P : \mathbb{B} \mapsto \mathbb{N} : b \rightarrow f_P(b)$ where $\mathbb{B}$

1. To be precise, the BPQs $b_{11}$, $b_{12}$, $b_{13}$, $b_{14}$, $b_{15}$, $b_{21}$ and $b_{22}$ respectively correspond to the inclusion of Rogers' host-parasitoid model (1972), the host-parasitoid model by Nicholson and Bailey (1935), Holling's predation model (1959), Thompson's host-parasitoid model (1929), the predation model by Lotka and Volterra (1925, 1926), a logistic population growth model (Verhulst, 1838) and an exponential population growth model (Malthus, 1798).

is the set of BPQs, $\mathbb{N}$ is the set of natural numbers and $f_P(b)$ equals the number of occurrences of $b$ in $b_1, \ldots, b_n$.

For example, let $P_{\text{model}}$ denote the OMP associated with the scenario model that contains three logistic population growth models ($b_{21}$), two Holling predation model ($b_{13}$) and one competition model ($b_{31}$). Therefore,

$$P_{\text{model}} = b_{21} \oplus b_{21} \oplus b_{21} \oplus b_{13} \oplus b_{13} \oplus b_{31}$$

and hence:

$$f_{P_{\text{model}}}(b) = \begin{cases} 3 & \text{if } b = b_{21} \\ 2 & \text{if } b = b_{13} \\ 1 & \text{if } b = b_{31} \\ 0 & \text{otherwise} \end{cases}$$

By describing OMPs as functions, the concept of combinations of OMPs becomes clear. For two OMPs $P_1$ and $P_2$, the combined preference $P_1 \oplus P_2$ is defined as:

$$f_{P_1 \oplus P_2} : \mathbb{B} \mapsto \mathbb{N} : b \rightarrow f_{P_1 \oplus P_2}(b) = f_{P_1}(b) + f_{P_2}(b)$$

Note that the combination operator $\oplus$ is assumed to be commutative, associative and strictly monotonic ($P \prec P \oplus P$). The latter assumption is made to better reflect the ideas underpinning conventional utility calculi (Binger & Hoffman, 1998).

### 2.2.3 Partial ordering of OMPs

Based on the combinations of OMPs, a partial order $\preccurlyeq$ over the OMPs can be computed by exploiting the constituent BPQs of the OMPs considered. This partial order implies that a comparison of any pair of OMPs either returns equal preference ($=$), smaller preference ($\prec$), greater preference ($\succ$) or incomparable preference ($?$). This calculus is developed assuming the following:

- *Prioritisation*: A combination of BPQs is never an order of magnitude greater than its constituent BPQs. That is, given the set of BPQs belonging to the same order of magnitude $\{b_1, b_2, \ldots, b_n\} \subseteq O_1$ and a BPQ $b \in O_2$ belonging to a higher order of magnitude, i.e. $O_1 \ll O_2$, then

  $$b_1 \oplus b_2 \oplus \ldots \oplus b_n \prec b$$

  With respect to the ongoing example, this means that any BPQ taken from the order of magnitude $O_1$ is preferred over any combination of BPQs taken from $O_2$. In other words, the choice of a model to describe a host-parasitoid phenomenon is considered more important than the choice of population growth model (see Figure 1).

  Prioritisation also means that distinctions at higher orders of magnitude are considered to be more significant than those at lower orders of magnitude. Consider a number of BPQs $b_1, \ldots, b_{m-1}, b_m, \ldots, b_n$ taken from one order of magnitude $O_1$ and a pair of BPQs $\{b, b'\}$ taken from an order of magnitude that is higher than $O_1$. If $b < b'$, then (irrespective of the ordering of the BPQs taken from $O_1$)

  $$b_1 \oplus \ldots \oplus b_{m-1} \oplus b \prec b_m \oplus \ldots \oplus b_n \oplus b'$$

- *Strict monotonicity*: Even though distinctions at higher orders of magnitude are more significant, distinctions at lower orders of magnitude are not negligible. That is, given an OMP $P$ and two BPQs $b_1$ and $b_2$ taken from the same order of magnitude with $b_1 < b_2$, then (irrespective of the orders of magnitude of the BPQs that constitute $P$)

  $$b_1 \oplus P \prec b_2 \oplus P$$

  For instance, the preference ordering depicted in Figure 1 shows that a scenario model with a Roger's host-parasitoid model and two logistic predation models is preferred over one with a Roger's host-parasitoid model and two exponential predation models:

  $$b_{11} \oplus b_{22} \oplus b_{22} \prec b_{11} \oplus b_{21} \oplus b_{21}$$

  Note that this is a departure from conventional order-of-magnitude reasoning. If the OMPs associated with two (partial) outcomes contain equal BPQs at a higher order of magnitude, it is usually desirable to compare both solutions further in terms of the (less important) constituent BPQs at lower orders of magnitude, as the example illustrated. However, conventional order-of-magnitude reasoning techniques can not handle this.

- *Partial ordering maintenance*: Conventional order-of-magnitude reasoning is motivated by the need for abstract descriptions of real-world behaviour, whereas the OMP calculus is motivated by incomplete knowledge for decision making. As opposed to conventional order-of-magnitude reasoning and real numbers, OMPs are not necessarily totally ordered. This implies that, when the user states, for example, that $b_1 < b_2 < b$ and that $b_3 < b_4 < b$, the explicit absence of ordering information between the BPQs in $\{b_1, b_2\}$ and those in $\{b_3, b_4\}$ means that the user is unable to compare them (e.g. because they are entirely different things). Consequently, $b_1 \oplus b_2$ would be deemed incomparable to $b_3 \oplus b_4$ (i.e. $b_1 \oplus b_2 ? b_3 \oplus b_4$), rather than roughly equivalent.

From the above, it can be derived that given two OMPs $P_1$ and $P_2$ and an order of magnitude $O$, $P_1$ is *less or equally preferred to* $P_2$ *with respect to the order of magnitude* $O$ (denoted $P_1 \preccurlyeq_O P_2$) provided that

$$\forall b_i \in O, \big(f_{P_1}(b_i) + \sum_{b_j \in O, b_i < b_j} f_{P_1}(b_j)\big) \leq \big(f_{P_2}(b_i) + \sum_{b_j \in O, b_i < b_j} f_{P_2}(b_j)\big)$$

Thus, comparing two OMPs within an order of magnitude can yield four possible results:

- $P_1$ is less preferred than $P_2$ with respect to $O$ ($P_1 \prec_O P_2$) iff $(P_1 \preccurlyeq_O P_2) \wedge \neg(P_2 \preccurlyeq P_1)$,
- $P_1$ is more preferred than $P_2$ with respect to $O$ ($P_1 \succ_O P_2$) iff $\neg(P_1 \preccurlyeq_O P_2) \wedge (P_2 \preccurlyeq P_1)$,
- $P_1$ is equally preferred than $P_2$ with respect to $O$ ($P_1 =_O P_2$) iff $(P_1 \preccurlyeq_O P_2) \wedge (P_2 \preccurlyeq P_1)$, and
- $P_1$ is incomparable to $P_2$ with respect to $O$ ($P_1 ?_O P_2$) iff $\neg(P_1 \preccurlyeq_O P_2) \wedge \neg(P_2 \preccurlyeq P_1)$.

In the ongoing example of Figure 1, for instance, the preference of a scenario model with a Roger's host-parasitoid model and a Holling predation model is $P_1 = b_{11} \oplus b_{13}$ and the preference of a scenario model with a Roger's host-parasitoid model and a Lotka-Volterra predation model is $P_2 = b_{11} \oplus b_{15}$. The latter model is less than or equally preferred to the former within the "host-parasitoid" order of magnitude ($O_1$), i.e. $P_2 \preccurlyeq_{O_1} P_1$, because

$$\begin{aligned}
f_{P_2}(b_{11}) = 1 \leq 1 = f_{P_1}(b_{11}),\\
f_{P_2}(b_{11}) \oplus f_{P_2}(b_{12}) = 1 \leq 1 = f_{P_1}(b_{11}) \oplus f_{P_1}(b_{12}),\\
f_{P_2}(b_{11}) \oplus f_{P_2}(b_{13}) = 1 \leq 2 = f_{P_1}(b_{11}) \oplus f_{P_1}(b_{13}),\\
f_{P_2}(b_{11}) \oplus f_{P_2}(b_{12}) \oplus f_{P_2}(b_{14}) = 1 \leq 1 = f_{P_1}(b_{11}) \oplus f_{P_1}(b_{12}) \oplus f_{P_1}(b_{14}),\\
f_{P_2}(b_{11}) \oplus f_{P_2}(b_{12}) \oplus f_{P_2}(b_{13}) \oplus f_{P_2}(b_{14}) = 2 \leq 2 = f_{P_1}(b_{11}) \oplus f_{P_1}(b_{12}) \oplus f_{P_1}(b_{13}) \oplus f_{P_1}(b_{14}).
\end{aligned}$$

Similarly, it can be established that the reverse, i.e. $P_1 \preccurlyeq_{O_1} P_2$, is not true. Therefore, the latter scenario model is less preferred than the former within $O_1$, i.e. $P_2 \prec_{O_1} P_1$.

The above result can be further generalised such that given two OMPs $P_1$ and $P_2$, $P_1$ is *less or equally preferred to* $P_2$ (denoted $P_1 \preccurlyeq P_2$) if

$$\forall O_i \in \mathbf{O}, (P_1 \preccurlyeq_{O_i} P_2) \vee (\exists O_j \in \mathbf{O}, O_i \ll O_j \wedge P_1 \prec_{O_j} P_2)$$

More generally, the relations $\prec$, $\succ$, $=$ and $?$ can be derived in the same manner as with the relation $\preccurlyeq$ where $\prec_O$, $\succ_O$, $=_O$ and $?_O$ with $\preccurlyeq_O$.

To illustrate the utility of such orderings, consider the scenario of one predator population that feeds on two prey populations while the two prey populations compete for scarce resources. The following are two plausible scenario models for this scenario:

- Model 1 contains two Holling predation models and three logistic population growth models, and has preference $P_1 = b_{13} \oplus b_{13} \oplus b_{21} \oplus b_{21} \oplus b_{21}$.
- Model 2 contains one competition model, two Holling predation models, two logistic population growth models and an exponential population growth model, and has preference $P_2 = b_{13} \oplus b_{13} \oplus b_{21} \oplus b_{21} \oplus b_{22} \oplus b_{31}$.

As demonstrated earlier, it can be shown that $P_1 =_{O_1} P_2$, $P_1 \succ_{O_2} P_2$, and $P_1 \prec_{O_3} P_2$. From these relations it follows that $P_1 \preccurlyeq P_2$ because

- for $O_1$: $P_1 \preccurlyeq_{O_1} P_2$ since $P_1 =_{O_1} P_2$,
- for $O_2$: there exists an order of magnitude $O_3$ where $O_3 \gg O_2$ and $P_1 \prec_{O_3} P_2$,
- for $O_3$: $P_1 \preccurlyeq_{O_3} P_2$ since $P_1 \prec_{O_3} P_2$.

As the reverse is not true, it can be concluded that scenario model 2 is preferred over scenario model 1.

### 2.3 Solving aDPCSPs

This section presents a basic algorithm for solving aDPCSPs. Although OMPs are used in this work, this algorithm can take any aDPCSP provided that it employs a preference calculus with a commutative, associative and monotonic combination operator. However, the use of OMPs provides a convenient way of specifying incomplete preference information.

An aDPCSP is similar to valued CSPs as presented by Schiex, Fargier and Verfaillie (1995) and also to semiring based CSPs (Bistarelli, Montanari, & Rossi, 1997). However, it extends both approaches with activity constraints and involves different underlying presumptions in its valuation structure. The preference valuations in this work are allowed to be ordered partially, as opposed to the valued CSPs.

An aDPCSP represents an important type of constraint satisfaction optimisation problem (Tsang, 1993). In order to tackle the optimisation of preferences an A* type algorithm is employed (Hart, Nilsson, & Raphael, 1968; Raphael, 1990). A* algorithms are known to be efficient in terms of the total number of nodes explored in an effort to find optimal solutions, with a given amount of information. On the downside, they have an exponential space complexity. Naturally, a number of alternative approaches could have been explored, including conventional constraint-based solving methods such as depth first branch and bound search. However, the use of an A*-like algorithm is sufficient for solving the aDPCSPs in the domain of the present interest. In particular, algorithm 1 implements an A* search strategy that is capable of handling activity constraints, which involves the use of basic CSP techniques such as constraint propagation and backtracking.

An A* algorithm maintains the explored attribute-value assignments by means of a priority queue $Q$ of nodes. Each node $n$ in $Q$ corresponds to a set of attribute-value assignments: solution$(n)$. The search proceeds through a number of iterations. At each iteration, a node $n$ is taken from $Q$, and replaced by nodes that extend solution$(n)$ with an additional attribute-value assignment. More specifically, for each node $n$ in $Q$, a set $X_u(n)$ of remaining active but unassigned attributes is maintained. At each iteration, the possible assignments of the first attribute $x \in X_u(n)$, where $n$ is the node taken from $Q$ at the current iteration, are processed. For every assignment $x : d$ that is consistent with solution$(n)$ (i.e. solution$(n) \cup \{x : d\}, \mathbf{C} \nvdash \bot$), a new child node $n'$, with solution$(n')$ = solution$(n) \cup \{x : d\}$ and $X_u(n') = X_u(n) - \{x\}$, is created and added to $Q$.

The activity constraints are processed via propagation rather than constraint satisfaction. Whenever a node $n$ is taken from $Q$ such that $X_u(n)$ is empty, the activity constraints are fired in order to obtain a new set of active but unassigned attributes. That is, $X_u(n)$ is assigned

$$\{x_i \mid \text{solution}(n), \mathbf{A} \vdash \text{active}(x_i)\} - X_a(n)$$

where $X_a(n)$ represents the active, but already assigned attributes in node $n$.

In the priority queue $Q$, nodes are maintained by means of two heuristics: committed preference $CP(n)$ and potential preference $PP(n)$. Here, given a node $n$,

$$CP(n) = \oplus_{x:d \in \text{solution}(n)} P(x : d)$$
$$PP(n) = CP(n) \oplus (\oplus_{x \in X_{nd}(n)} \max_{d \in D_x} P(x : d))$$

where $X_{nd}(n)$ is the set of unassigned attributes that can still be activated given the partial assignment solution($n$) (as indicated previously, the actual implementation employs an assumption-based truth maintenance system (de Kleer, 1986) to efficiently determine which attribute's activity can no longer be supported). In other words, $CP(n)$ is the preference associated with the partial attribute-value assignment in node $n$ and $PP(n)$ is $CP(n)$ combined with the highest possible preference assignments taken from all the values of the domains of those attributes in $X_{nd}(n)$. Thus, $PP(n)$

**Algorithm 1:** $\textsc{solve}(\mathbf{X}, \mathbf{D}, \mathbf{C}, \mathbf{A}, P)$

$n \leftarrow$ new node;
solution$(n) \leftarrow \{\}$;
$X_u(n) \leftarrow \{x_i \mid \{\}, \mathbf{A} \vdash \text{active}(x_i)\}$;
$X_a(n) \leftarrow \{\}$;
$CP(n) \leftarrow 0$;
$PP(n) \leftarrow \oplus_{x \in \mathbf{X}} \max_{d \in D(x)} P(x : d)$;
$Q \leftarrow$ createOrderedQueue$()$;
enqueue$(Q, n, PP(n), CP(n))$; **while** $Q \neq \emptyset$
**do**
  $n \leftarrow$ dequeue$(Q)$;
  **if** $X_u(n) \neq \emptyset$
  **then**
    $x \leftarrow$ first$(X_u(n))$;
    $\textsc{process}(x, n, \mathbf{C}, \mathbf{A}, P, Q)$;
  **else**
    $X_u(n) \leftarrow \{x_i \mid \text{solution}(n), \mathbf{A} \vdash \text{active}(x_i)\} - X_a(n)$;
    **if** $X_u(n) = \emptyset$
    **then**
      $n_{\text{next}} \leftarrow$ first$(Q)$;
      **if** $CP(n) \not\prec PP(n_{\text{first}})$
      **then return** (solution$(n)$);
      **else**
        $PP(n) \leftarrow CP(n)$;
        enqueue$(Q, n, PP(n), CP(n))$;
    **else**
      $x \leftarrow$ first$(X_u(n))$;
      $\textsc{process}(x, n, \mathbf{C}, \mathbf{A}, P, Q)$;

**procedure** $\textsc{process}(x, n_{\text{parent}}, \mathbf{C}, \mathbf{A}, P, Q)$
**for** $d \in D(x)$
**do**
  **if** solution$(n_{\text{parent}}) \cup \{x : d\}, \mathbf{C} \nvdash \bot$
  **then**
    $n_{\text{child}} \leftarrow$ new node;
    solution$(n_{\text{child}}) \leftarrow$ solution$(n_{\text{parent}}) \cup \{x : d\}$;
    $X_d \leftarrow$ deactivated(solution$(n_{\text{child}}), X(n_{\text{parent}}))$;
    $X_{nd}(n_{\text{child}}) \leftarrow X_{nd}(n_{\text{parent}}) - \{x\} - X_d$;
    $X_a(n_{\text{child}}) \leftarrow X_a(n_{\text{parent}}) \cup \{x\}$;
    $X_u(n_{\text{child}}) \leftarrow X_u(n_{\text{parent}}) - \{x\}$;
    $CP(n_{\text{child}}) \leftarrow CP(n_{\text{parent}}) \oplus P(x : d)$;
    $PP(n_{\text{child}}) \leftarrow CP(n_{\text{child}}) \oplus \oplus_{x \in X_{nd}(n)} \max_{d \in D(x)} P(x : d)$;
    enqueue$(Q, n_{\text{child}}, PP(n_{\text{child}}), CP(n_{\text{child}}))$;

computes an upper boundary on the preference of an aDPCSP solution that includes the partial attribute-value assignments corresponding to $n$.

The following theorem shows that algorithm 1 is guaranteed to find the set of attribute-value pairs with the highest combined preferences, within the set of solutions that satisfy the constraints.

**Theorem 1** $\textsc{solve}(\mathbf{X}, \mathbf{D}, \mathbf{C}, \mathbf{A}, P)$ *is admissible*

Proof*:* $\textsc{solve}(\mathbf{X}, \mathbf{D}, \mathbf{C}, \mathbf{A}, P)$ *is an A* algorithm guided by a heuristic function* $PP(n) = CP(n) \oplus h(n)$*, where* $CP(n)$ *is the actual preference of node* $n$ *and* $h(n) = \oplus_{x \in X_{nd}(n)} \max_{d \in D_x} P(x : d)$*. It follows from the previous discussion that* $h(n)$ *is greater than or equal to the combined preference of any value-assignment of unassigned attributes that is consistent with the partial solution of* $n$*. In this algorithm, the nodes* $n$ *are maintained in a priority queue in descending order of* $PP(n)$*. Let* $\delta$ *be a distance function that reverses the preference ordering such that* $\delta(P_1) \prec \delta(P_2) \leftrightarrow P_1 \succ P_2$*.* $\textsc{solve}(\mathbf{X}, \mathbf{D}, \mathbf{C}, \mathbf{A}, P)$ *can then be described as an A* algorithm, where the nodes* $n$ *in the priority* *queue $Q$ are ordered in ascending order of $\delta(PP(n))$, such that $\delta(PP(n)) = \delta(CP(n)) \oplus \delta(h(n))$ and $\delta(h(n))$ is a lower bound on the distance between $n$ and the optimal solution. Therefore, following the work by Hart, Nilsson and Raphael (1968),* $\textsc{solve}(\mathbf{X}, \mathbf{D}, \mathbf{C}, \mathbf{A}, P)$ *is an admissible algorithm, guaranteed to find a solution $S$ with a minimal $\delta(P(S))$ or a maximal $P(S)$.*

To illustrate algorithm 1, consider the problem of finding an ecological model that describes the behaviour of two populations, one of which predates on the other. An aDPCSP is constructed for the compositional modelling problem with the following attributes and domains. Note that section 3 demonstrates how the attributes, domains and constraints of this problem can be constructed automatically and that section 4 illustrates these ideas in the context of a larger example.

$$\begin{aligned}
\mathbf{X} &= \{x_1, x_2, x_3, x_4, x_5, x_6\} \\
D_{x_1} &= \{\text{yes}, \text{no}\} \\
D_{x_2} &= \{\text{yes}, \text{no}\} \\
D_{x_3} &= \{\text{yes}, \text{no}\} \\
D_{x_4} &= \{\text{other}, \text{logistic}\} \\
D_{x_5} &= \{\text{other}, \text{logistic}\} \\
D_{x_6} &= \{\text{Holling}, \text{Lotka-Volterra}\}
\end{aligned}$$

The attributes $x_1$, $x_2$ and $x_3$ respectvely describe the relevance of the following phenomena: the change in size of the predator population, the change in size of the prey population and the predation of the prey by the predator. The attributes $x_4$ and $x_5$ represent the choice of type of population growth model. Two types of such models are incorporated in the problem: the logistic one and the "other". Finally, attribute $x_6$ is associated with the choice of model type of the predation phenomenon. Here, two types of model, the Holling model and the Lotka-Volterra model, are included.

Because the Holling predation model presumes that logistic models are employed to describe population growth, and because the Lotka-Volterra Model incorporates its own population growth model, the combinations of assignments to $x_4$, $x_5$, and $x_6$ are restricted. Hence, the aDPCSP contains a set $\mathbf{C} = \{c_{\{x_4,x_6\}}, c_{\{x_5,x_6\}}\}$ of compatibility constraints, with:

$$\begin{aligned}
c_{\{x_4,x_6\}} &= \{\langle x_4 : \text{other}, x_6 : \text{Lotka-Volterra}\rangle, \langle x_4 : \text{logistic}, x_6 : \text{Holling}\rangle\} \\
c_{\{x_5,x_6\}} &= \{\langle x_5 : \text{other}, x_6 : \text{Lotka-Volterra}\rangle, \langle x_5 : \text{logistic}, x_6 : \text{Holling}\rangle\}
\end{aligned}$$

Furthermore, a model type of predator/prey growth must be selected if and only if the corresponding population growth phenomenon is deemed relevant. Also, a model type of predation must be selected if and only if both population growth phenomena and the predation phenomenon are deemed relevant (because ecological models describing predation rely on submodels describing population growth of the predator and the prey). Hence, the aDPCSP contains a set $\mathbf{A} = \{a_{x_4,\{x_1\}}, a_{x_5,\{x_2\}}, a_{x_6,\{x_1,x_2,x_3\}}\}$ of activity constraints, with:

$$
\begin{aligned}
a_{x_4,\{x_1\}} &= \{\langle x_1 : \text{yes}, \text{active}(x_4)\rangle, \langle x_1 : \text{no}, \neg\text{active}(x_4)\rangle\} \\
a_{x_5,\{x_2\}} &= \{\langle x_2 : \text{yes}, \text{active}(x_5)\rangle, \langle x_2 : \text{no}, \neg\text{active}(x_5)\rangle\} \\
a_{x_6,\{x_1,x_2,x_3\}} &= \{\langle x_1 : \text{yes}, x_2 : \text{yes}, x_3 : \text{yes}, \text{active}(x_4)\rangle, \langle x_1 : \text{yes}, x_2 : \text{yes}, x_3 : \text{no}, \neg\text{active}(x_4)\rangle, \\
&\quad \langle x_1 : \text{yes}, x_2 : \text{no}, x_3 : \text{yes}, \neg\text{active}(x_4)\rangle, \langle x_1 : \text{yes}, x_2 : \text{no}, x_3 : \text{no}, \neg\text{active}(x_4)\rangle, \\
&\quad \langle x_1 : \text{no}, x_2 : \text{yes}, x_3 : \text{yes}, \neg\text{active}(x_4)\rangle, \langle x_1 : \text{no}, x_2 : \text{yes}, x_3 : \text{no}, \neg\text{active}(x_4)\rangle, \\
&\quad \langle x_1 : \text{no}, x_2 : \text{no}, x_3 : \text{yes}, \neg\text{active}(x_4)\rangle, \langle x_1 : \text{no}, x_2 : \text{no}, x_3 : \text{no}, \neg\text{active}(x_4)\rangle\}
\end{aligned}
$$

Finally, let the preference calculus consist of two orders of magnitude $O_{\text{growth}}$ and $O_{\text{predation}}$, with $O_{\text{growth}} \ll O_{\text{predation}}$, where

$$
\begin{aligned}
O_{\text{growth}} =& \{p_{\text{other}}, p_{\text{logistic}}\} \text{ with } p_{\text{logistic}} < p_{\text{other}} \\
O_{\text{predation}} =& \{p_{\text{Holling}}, p_{\text{Lotka-Volterra}}\} \text{ with } p_{\text{Lotka-Volterra}} < p_{\text{Holling}}
\end{aligned}
$$

The OMP assignments are as follows:

$$
\begin{aligned}
P(x_4 : \text{other}) = P(x_5 : \text{other}) =& p_{\text{other}} \\
P(x_4 : \text{logistic}) = P(x_5 : \text{logistic}) =& p_{\text{logistic}} \\
P(x_6 : \text{Holling}) =& p_{\text{Holling}} \\
P(x_6 : \text{Lotka-Volterra}) =& p_{\text{Lotka-Volterra}}
\end{aligned}
$$

When applied to this problem, algorithm 1 initialises the search by creating a node $n_0$, where:

- $X_u(n_0)$, the set of currently active attributes, is initialised to $\{x_1, x_2, x_3\}$, because the activity of these attributes does not depend on other attribute-value assignments.
- $X_a(n_0)$ and $CP(n_0)$ are initialised to the empty set and to 0 respectively, since no attributes have been assigned yet.
- Finally, $PP(n_0)$ equals $p_{\text{other}} \oplus p_{\text{other}} \oplus p_{\text{Holling}}$ because this is the combination of highest OMPs associated with each domain.

This initial node is enqueued in $Q$. Next, the algorithm proceeds through a number of iterations. At each iteration, the node with most potential (as measured by $PP$ and $CP$) is dequeued, and its children are generated and enqueued in $Q$. The nodes that are created in this way are depicted in Figure 2. The number $i$ in the subscript of each node $n_i$ indicates the order of node generation, and the thick arrows show the order in which the search space is explored.

Note that there are three important features of the algorithm that could not be clearly demonstrated within Figure 2. Firstly, at node $n_5$, the initial set of unassigned attributes is exhausted: $X_u(n_5) = \{\}$. Therefore, the activity constraints are fired when $n_5$ is explored. Because $n_5$ corresponds to the assignment $\{x_1 : \text{yes}, x_2 : \text{yes}, x_3 : \text{yes}\}$, the remaining attributes are activated and $X_u(n_5)$ is reset to $\{x_4, x_5, x_6\}$.

Secondly, node $n_{12}$ corresponds to an assignment of all (active) attributes that is consistent with the activity and compatibility constraints:

$$\{x_1 : \text{yes}, x_2 : \text{yes}, x_3 : \text{yes}, x_4 : \text{other}, x_5 : \text{other}, x_6 : \text{Lotka-Volterra}\}$$

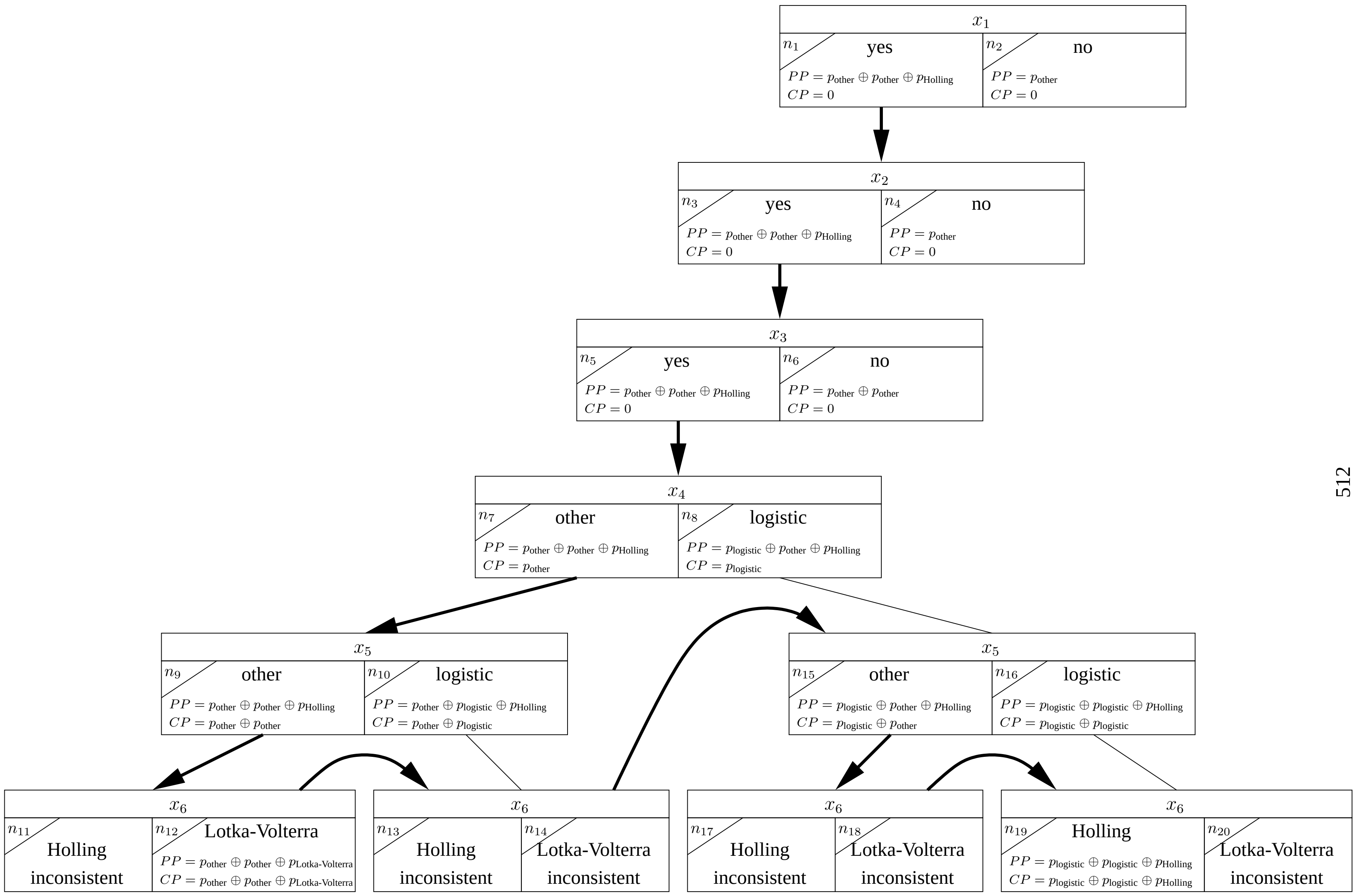

Figure 2: Search space explored by algorithm 1 when solving sample aDPCSP

This assignment is not a solution to the aDPCSP, because the corresponding preference is not guaranteed to be maximal (and, the assignment is, in fact, not optimal). After the creation of $n_{12}$, the priority queue $Q$ looks as follows (the ordering between $n_2$ and $n_4$ may vary since $PP(n_2) = PP(n_4)$ and $CP(n_2) = CP(n_4)$):

$$\{n_{10}, n_8, n_{12}, n_6, n_2, n_4\}$$

Therefore, the next node to be explored (after $n_9$ and the subsequent creation of $n_{12}$) is $n_{10}$.

Thirdly, node $n_{19}$ does correspond with an optimal solution. After its creation, $Q$ equals:

$$\{n_{19}, n_{12}, n_6, n_2, n_4\}$$

As a consequence, $n_{19}$ is dequeued in the next iteration. Because no children of $n_{19}$ can be created ($X_u(n_{19}) = \emptyset$ and the activity constraints activate no more attributes), $n_{19}$ is retained as a solution.

If the user is interested in finding multiple alternative solutions, the search may proceed until $Q$ contains no more nodes with a $PP$ value that is not smaller than the maximum preference of the first solution. In this case, $PP(n_{12}) \prec CP(n_{19})$ and hence, there is only one solution to this aDPCSP.

## 3. Compositional Model Repositories

The aDPCSPs discussed in the previous section provide the foundation for the development of the compositional model repositories. This section specifies the problem that a compositional model repository is built to solve and shows how it can be translated into an aDPCSP, and hence be resolved using the proposed aDPCSP solution algorithm.

### 3.1 Background: assumption based truth maintenance

An ATMS is a mechanism that keeps track of how each piece of inferred information depends on presumed information and facts and of how inconsistencies arise. In an ATMS, each piece of information used or derived by the problem solver is stored as a *node*. Certain pieces of information are not known to be true and cannot be inferred from other pieces of information, yet plausible inference may be drawn from them. Such nodes are categorised by a special type and referred to as *assumptions*.

Inferences between pieces of information are maintained within the ATMS as dependencies between the corresponding nodes. In its extended form (see de Kleer, 1988; or Keppens, 2002), the ATMS can take inferences, called *justifications* of the form $n_i \wedge \ldots \wedge n_j \wedge \neg n_k \wedge \ldots \wedge \neg n_l \rightarrow n_m$, where $n_i, \ldots, n_j, n_k, \ldots, n_l, n_m$ are nodes that the problem solver is interested in. An ATMS can also take a specific type of justification, called *nogood*, that leads to an inconsistency, of the form $n_i \wedge \ldots \wedge n_j \wedge \neg n_k \wedge \ldots \wedge \neg n_l \rightarrow \perp$ (meaning that at least one of the statements in $\{n_i, \ldots, n_j, \neg n_k, \ldots, \neg n_l\}$ must be false). In the ATMS, these nogoods are represented as justifications of a special node, called the *nogood node*.

Based on the given justifications and nogoods, the ATMS computes a *label* for each (non-assumption) node. A label is a set of *environments* and an environment is a set of assumptions. In particular, an environment $E$ depicts a possible world where all the assumptions in $E$ are true. Thus, the label $\mathcal{L}(n)$ of a node $n$ describes all possible worlds in which $n$ can be true. The label computation algorithm of the ATMS guarantees that each label is:

- *Sound* - All assumptions in any environment within the label of a node being true is a sufficient condition to derive that node:

$$\forall E \in \mathcal{L}(n), [(\wedge_{n_i \in E} n_i) \wedge (\wedge_{\neg n_i \in E} \neg n_i)] \vdash n$$

- *Consistent* - No environment in the label of a node, other than the nogood node, describes an impossible world:

$$\forall E \in \mathcal{L}(n), [(\wedge_{n_i \in E} n_i) \wedge (\wedge_{\neg n_i \in E} \neg n_i)] \nvdash \perp$$

- *Minimal* - The label does not contain possible worlds that are less general than one of the other possible worlds it contains (i.e. environments that are supersets of other environments in the label):

$$\forall E \in \mathcal{L}(n) \nexists E' \in \mathcal{L}(n), E' \subset E$$

- *Complete* - The label of each node, other than the nogood node, describes all possible worlds in which that node can be inferred:

$$\begin{aligned} \forall E, & [(\wedge_{n_i \in E} n_i) \wedge (\wedge_{\neg n_i \in E} \neg n_i) \vdash n] \\ & \exists E' \in \mathcal{L}(n), [(\wedge_{n_i \in E'} n_i) \wedge (\wedge_{\neg n_i \in E'} \neg n_i) \vdash n] \end{aligned}$$

### 3.2 Knowledge Representation

As with any other knowledge-based approach, building a compositional modeller requires a formalism for the specification of its inputs, its outputs and its knowledge base. The work developed here is loosely based on the compositional modelling language (Bobrow, Falkenhainer, Farquhar, Fikes, Forbus, Gruber, Iwasaki, & Kuipers, 1996), a proposed standard knowledge representation formalism for compositional modellers, but adapted to meet the challenges of the ecological compositional modelling problems identified in the introduction.

#### 3.2.1 Preliminary concepts

The most primitive constructs in a compositional modeller are participants, relations and assumptions. This subsection summarises these concepts and explains how they are represented herein.

*Participants*[2] refer to the objects of interest, which are involved in the scenario or its model. These participants may be real-world objects or conceptual objects, such as variables that express features of real-world objects in a mathematical model. For instance, a population of a species is a typical example of a real-world object, and a variable that expresses the number of individuals of this species forms an example of a conceptual object. It is natural to group objects that share something in common into classes. Participants are herein grouped into *participant classes*, with each representing a set of participants that share certain common features. Each class will be given a name for easy reference.

*Relations* describe how the participants are related to one another. As with participants, some relations represent a real-world relationship, such as:

2. Some of the previous work in compositional modelling refers to these as *individuals* and *quantities*, but such names would not suit the present application. Ecological models typically describe the behaviour of populations rather than that of individuals and it is often hard to distinguish between quantities.

$$\text{predation(frog, insect)} \tag{1}$$

Other relations may be conceptual in nature, such as equation (2), which describes an important textbook model of logistic population growth (Ford, 1999):

$$\frac{d}{dt}\text{change} = \text{parameter} \times \text{size} \times (1 - \frac{\text{size}}{\text{capacity}}) \tag{2}$$

To be consistent with other compositional modelling approaches, this paper employs a LISP-style notation for relations. As such, the above two sample relations become:

```
(predation frog insect)                                    (1)
(d/dt change (* change-rate size (- 1 (/ size capacity))))   (2)
```

*Assumptions* form a special type of relation that are employed to distinguish between alternative model design decisions. Internally, assumptions will be stored in the form of assumption nodes in the ATMS (see section 3.3.1), but in the knowledge base, assumptions appear as relations with a specific syntax and semantics.

Two *types* of assumptions are employed in this article. *Relevance assumptions* state what phenomena are to be included in or excluded from the scenario model. Typical examples of phenomena are the population growth and predation phenomena. The general format of a relevance assumption is shown in (3). The phenomenon that is incorporated in the scenario model when describing a relevance assumption is identified by ⟨*name*⟩ and is specific to the subsequent participants or relations. For example, relevance assumption (4) states that the growth of participant `?population` is to be included in the model.

`(relevant` ⟨*name*⟩ [{⟨*participant*⟩} | ⟨*relation*⟩]`)` (3)

`(relevant growth ?population)` (4)

*Model assumptions* specify which type of model is utilised to describe the behaviour of a certain participant or relation. Typical examples of model types include the exponential (Malthus, 1798) and the logistic (Verhulst, 1838) model types of population growth. The formal specification of a model assumption is given in (5). Often the ⟨*name*⟩ in (5) corresponds to the name of a known (partial) model of the phenomenon or process being described. The example in (6) states that the population `?population` is being modelled using the logistic approach.

`(model` [⟨*participant*⟩ | ⟨*relation*⟩] ⟨*name*⟩`)` (5)

`(model ?population logistic)` (6)

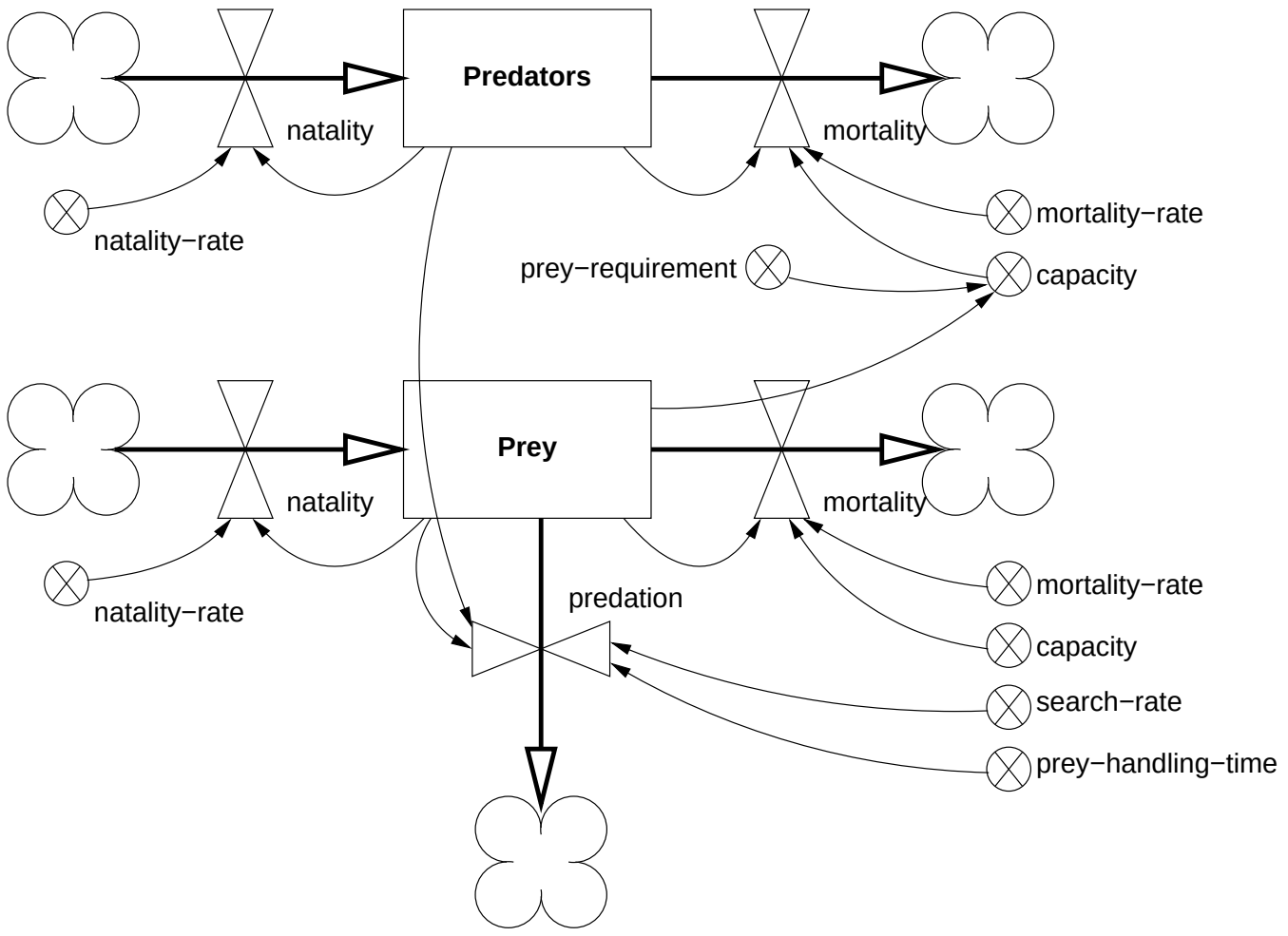

Figure 3: Stock flow diagram of `predator prey` scenario model

### 3.2.2 Scenarios and scenario models

As formalised by Keppens and Shen (2001b), a compositional modeller takes two inputs and produces one output. The first input is a representation (which is itself a model) that describes the system of interest by means of an accessible formalism. This model, which normally consists of (mainly) real-world participants and their interrelationships, is called the *scenario*. The second input is the *task description*. It is a formal description of the criteria by which the adequacy of the output is evaluated. The output is a new model that describes the scenario in a more detailed formalism, usually a set of variables and equations, which the model-based reasoner can employ readily. Such a model, which normally contains conceptual participants and interrelationships, is called a *scenario model*. The aim of any compositional modeller is to translate the scenario into a scenario model, by means of the task description.

In this work, a model is formally defined by a tuple $\langle P, R\rangle$, where $P$ is a set of participants and $R$ is a set of relations over the participants in $P$. This definition applies to both the scenario and the scenario model. A typical example of a scenario is a description of a predator population, a prey population and a predation relation between the predator and the prey. This scenario is a model $\langle P, R\rangle$ with:

$$P = \{\texttt{predator}, \texttt{prey}\}$$
$$R = \{\texttt{(predation predator prey)}\}$$

The aim of the compositional model repository is to translate a scenario into a scenario model. Within this work, both systems dynamics stock-flow formalism (Forrester, 1968) and ordinary differential equations (ODEs) will be employed as the modelling formalisms. For example, a scenario model that corresponds to the above scenario is depicted in Figure 3. Formally, a scenario model is another model $\langle P, R\rangle$ and in this case

$$P = \{N_{\text{predator}}, B_{\text{predator}}, D_{\text{predator}}, N_{\text{prey}}, B_{\text{prey}}, D_{\text{prey}}, P_{\text{prey}},\\ b_{\text{predator}}, b_{\text{prey}}, d_{\text{predator}}, d_{\text{prey}}, C_{\text{predator}}, C_{\text{prey}},\\ s_{(\text{prey},\text{predator})}, t_{(\text{prey},\text{predator})}, r_{(\text{predator},\text{prey})}\}$$

| Symbol | Variable name |
|---|---|
| $N_{\text{predator}}$, $N_{\text{prey}}$ | number of predators, prey |
| $B_{\text{predator}}$, $B_{\text{prey}}$ | natality of predators, prey |
| $D_{\text{predator}}$, $D_{\text{prey}}$ | mortality of predators, prey |
| $P_{\text{prey}}$ | predation of prey |
| $b_{\text{predator}}$, $b_{\text{prey}}$ | natality-rate of predators, prey |
| $d_{\text{predator}}$, $d_{\text{prey}}$ | mortality-rate of predators, prey |
| $C_{\text{predator}}$, $C_{\text{prey}}$ | capacity of predators, prey |
| $s_{(\text{prey,predator})}$ | search-rate |
| $t_{(\text{prey,predator})}$ | prey-handling-time |
| $r_{(\text{predator,prey})}$ | prey-requirement |

Table 1: Variables in the stock flow diagram and the mathematical model

$$
\begin{aligned}
R = \{ & \frac{d}{dt} N_{\text{predator}} = B_{\text{predator}} - D_{\text{predator}}, \\
& \frac{d}{dt} N_{\text{prey}} = B_{\text{prey}} - D_{\text{prey}} - P_{\text{prey}}, \\
& B_{\text{predator}} = b_{\text{predator}} \times N_{\text{predator}}, \\
& B_{\text{prey}} = b_{\text{prey}} \times N_{\text{prey}}, \\
& D_{\text{predator}} = d_{\text{predator}} \times N_{\text{predator}} \times \frac{N_{\text{predator}}}{C_{\text{predator}}}, \\
& D_{\text{prey}} = d_{\text{prey}} \times N_{\text{prey}} \times \frac{N_{\text{prey}}}{C_{\text{prey}}}, \\
& P_{\text{prey}} = \frac{s_{(\text{prey,predator})} \times N_{\text{prey}} \times N_{\text{predator}}}{1 + s_{(\text{prey,predator})} \times N_{\text{prey}} \times t_{(\text{prey,predator})}}, \\
& C_{\text{predator}} = r_{(\text{predator,prey})} \times N_{\text{prey}}, \\
& C_{\text{prey}} = N_{\text{prey}} \}
\end{aligned}
$$

The relation between the variables of the mathematical model and those used in the stock-flow diagram is given in table 1. Generally speaking, stock-flow diagrams are graphical representations of systems of (ordinary or qualitative) differential equations. In the automated modelling literature in general, and engineering and physical systems modelling in particular, more sophisticated representational formalisms have been developed to enable the identification of mathematical models of the behaviour of dynamic systems from observations. Examples include bond graphs (Karnopp, Margolis, & Rosenberg, 1990) and generalised physical networks (Easley & Bradley, 1999). However, the potential benefits of these more advanced formalisms are not exploited here, but remain as an interesting future work. Instead, stock-flow diagrams are employed throughout this paper as they are far more commonly used in ecological modelling (Ford, 1999).

It is often possible to construct multiple scenario models from a single given scenario, and the task specification is employed to guide the search for the most appropriate one(s). In this work, scenario models are selected on the basis of hard constraints and user preferences. The hard constraints stem from restrictions imposed on compositionality by the representational framework (see section 3.2.3) and from properties the scenario model is required to satisfy (see section 3.2.3). The user preferences express the user's subjective view as to which modelling approaches are more appropriate in the context of the current scenario (see section 2.2).

| Name | Syntax (infix notation) | Syntax (prefix notation) |
|---|---|---|
| Addition | ?var $= C^{+}$(formula) | `(== ?var (C-add formula))` |
| | ?var $= C^{-}$(formula) | `(== ?var (C-sub formula))` |
| Multiplication | ?var $= C^{\times}$(formula) | `(== ?var (C-mul formula))` |
| | ?var $= C^{\div}$(formula) | `(== ?var (C-div formula))` |
| Selection | ?var $= C^{\text{if},p}$(antecedent, formula) | `(== ?var (C-if antecedent formula :priority p))` |
| | ?var $= C^{\text{else}}$(formula) | `(== ?var (C-else formula)` |

Table 2: Composable functors and composable relations

### 3.2.3 THE KNOWLEDGE BASE

To construct scenario models from a given scenario, a compositional modeller relies on the use of a knowledge base that is particular to the problem domain. To illustrate the ideas, this section presents the constructs employed in the compositional modeller that is developed to synthesise scenario models in the ecological domain.

**Composable relations** The knowledge base in this approach consists of partial models that can be instantiated and composed into more complex scenario models. The composition of partial models into a scenario model may involve the composition of partial relations (coming from different partial models) in compounded relations. In the sample scenario model of section 3.2.2, the following relation describes the changes of population size of the prey population

$$\frac{d}{dt}N_{\text{prey}} = B_{\text{prey}} - D_{\text{prey}} - P_{\text{prey}} \tag{7}$$

In (7), $N_{\text{prey}}$ is the population size, $B_{\text{prey}}$ the number of births, $D_{\text{prey}}$ the number of natural deaths and $P_{\text{prey}}$ the number of prey who died due to predation. Thus, relation (7) actually describes two phenomena that affect the population size $N_{\text{prey}}$: natural population growth ($B_{\text{prey}} - D_{\text{prey}}$) and predation related deaths ($P_{\text{prey}}$). When constructing the knowledge base, it is desirable to represent these two phenomena in isolation because they do not always occur in combination. For example, some species do not have predators, and it is therefore unnecessary to always include predation as a cause of death. From this viewpoint, relation (7) can be seen as composed from different *composable relations* in the knowledge base:

$$\frac{d}{dt}N_{\text{prey}} = C^{+}(B_{\text{prey}}) \qquad \frac{d}{dt}N_{\text{prey}} = C^{-}(D_{\text{prey}}) \qquad \frac{d}{dt}N_{\text{prey}} = C^{-}(P_{\text{prey}})$$

The use of composable relations enables the knowledge base to cover as many combinations of the phenomena that may affect a relation as possible, by representing each phenomenon individually rather than precompiling everything together. Because only the component parts (i.e. the composable relations) of relations need to be represented, instead of all possible, and however complex, combinations of them, the knowledge base can be smaller and more effective. This section describes how such composable relations are represented in the knowledge base, as well as whether and how they can be composed to form compounded relations.

Composable relations are those containing composable functors and for which a method of composition exists (that describes how a complete set of composable relations can be composed). The composable functors employed are those proposed by Bobrow et al. (1996) with a new addition: composable selection. A summary of such composable relations is presented in table 2.

The composable relations introduced by Bobrow et al. (1996) are easy to understand. The formulae $f$ in $v = C^{+}(f)$ and $v = C^{-}(f)$ represent terms (respectively $f$ and $-f$) of a sum, and the formulae $f$ in $v = C^{\times}(f)$ and $v = C^{\div}(f)$ represent factors (respectively $f$ and $\frac{1}{f}$) of a product.

However, ecological models in use typically contain selection statements which declare that one certain equation must be employed when a condition is satisfied and some other one otherwise. Formally, a selection is a relation of the form

$$\text{if } c_1 \text{ then } v = r_1 \text{ else if } c_2 \ldots \text{ else } v = r_n \tag{8}$$

where $v$ is a participant, each $c_i$ (with $i = 1, \ldots, n-1$) is a relation describing a condition statement and each $r_j$ (with $j = 1, \ldots, n$) is a relation. This selection relation consists of the partial relations:

$$\begin{aligned} &\text{if } c_i \text{ then } v = r_i \qquad \text{with } i = 1, \ldots, n-1 \\ &\text{else } v = r_n \end{aligned}$$

Therefore, a selection relation can be composed from two types of composable relation. The first is a composable "if" relation, which has the form $v = C^{\text{if},p}(a, f)$, where $v$ is a participant, $p$ is an element taken from a total order, such as the set of natural numbers $\mathbb{N}$, which denotes the priority of the composable "if" relation in the sequence, and $a$ and $f$ are two given relations. The second type of composable relation is a composable "else" relation, which has the form $v = C^{\text{else}}(f_{\text{else}})$, where $f_{\text{else}}$ is a given relation assigned to $v$ if none of the antecedents in the composable "if" relations is true.

To illustrate this notation, the selection relation (8) can be composed from the following composable relations:

$$\begin{aligned} v &= C^{\text{if},p_1}(c_1, r_1) \\ &\vdots \\ v &= C^{\text{if},p_{n-1}}(c_{n-1}, r_{n-1}) \\ v &= C^{\text{else}}(r_n) \end{aligned}$$

with $p_1 > \ldots > p_{n-1}$.

To combine the composable relations, a number of rules are defined to implement the semantics of the representational formalism. In theory, a set of rules can be generated that enables the aggregation of any set of composable relations. In practice, however, a trade-off must be made between flexibility (the ability to combine many different types of composable relation) and comprehensibility (the use of a set of rules that is easily understood by the knowledge engineer who employs composable relations). Thus, the types of composable relations that can be combined has to be restricted.

Table 3 summarises what composable relations can be joined to form compounded relations. The principle guiding the construction of this table is to allow only the composition of relations of certain types for which a resulting compound relation is intuitively obvious. For example, according to Table 3, a composable addition relation $x = C^{+}(y)$ can be combined with a composable subtraction relation $x = C^{-}(z)$ because their combination is clearly $x = y - z$. However, according to Table 3, a composable addition relation $x = C^{+}(y)$ can not be combined with a composable multiplication relation $x = C^{\times}(z)$, because an arbitrary and non-intuitive rule would otherwise have to be defined to decide whether the compound relation would be $x = y + z$ or $x = y \times z$.

| | $C^{+}(f_2)$ | $C^{-}(f_2)$ | $C^{\times}(f_2)$ | $C^{\div}(f_2)$ | $C^{\text{if},p_2}(a_2, f_2)$ | $C^{\text{else}}(f_2)$ |
|---|---|---|---|---|---|---|
| $C^{+}(f_1)$ | yes | yes | no | no | no | no |
| $C^{-}(f_1)$ | yes | yes | no | no | no | no |
| $C^{\times}(f_1)$ | no | no | yes | yes | no | no |
| $C^{\div}(f_1)$ | no | no | yes | yes | no | no |
| $C^{\text{if},p_1}(a_1, f_1)$ | no | no | no | no | yes | yes |
| $C^{\text{if},p_2}(a_1, f_1)$ | no | no | no | no | no | yes |
| $C^{\text{else}}(f_1)$ | no | no | no | no | yes | no |

Table 3: Composibility of composable relations

The order in which the composable selections must be considered is defined by the priorities (or is implicit in the case of $C^{\text{else}}$). Therefore, composable selections can be combined with one another provided no two composable "if" relations have the same priority.

In order to derive the actual rules of composition, the sets of all composable relations with the same functor for a given model $\langle P, R \rangle$ are defined first:

$$
\begin{aligned}
R(v, C^{+}) &= \{v = C^{+}(f_i) \mid (v = C^{+}(f_i)) \in R\} \\
R(v, C^{-}) &= \{v = C^{-}(f_i) \mid (v = C^{-}(f_i)) \in R\} \\
R(v, C^{\times}) &= \{v = C^{\times}(f_i) \mid (v = C^{\times}(f_i)) \in R\} \\
R(v, C^{\div}) &= \{v = C^{\div}(f_i) \mid (v = C^{\div}(f_i)) \in R\} \\
R(v, C^{\text{if}}) &= \{v = C^{\text{if},p_i}(a_i, f_i) \mid (v = C^{\text{if},p_i}(a_i, f_i)) \in R\} \\
R(v, C^{\text{else}}) &= \{v = C^{\text{else}}(f_i) \mid (v = C^{\text{else}}(f_i)) \in R\}
\end{aligned}
$$

From this, the rules of composition can be built as given in the expressions (9), (10) and (11). They jointly state how a given set of composable relations can be rewritten as a single compound relation. Each of these rules contains a complete set of all composable relations in the antecedent. In particular, the antecedent of rule (9) contains the set of all composable addition and subtraction relations with the same participant $v$ in the left-hand side.

Similarly, the antecedent rule (10) contains the complete set of composable multiplication relations. Finally, the antecedent of rule (11) is satisfied for the complete set of composable if and else relations with the same left-hand participant $v$, provided that the priorities are strictly ordered (i.e. no two priorities are equal) and that there is only a single composable else relation. The latter two conditions are added because two composable if relations with the same priority or two composable else relations can not be compounded. The consequents of the rules of composition explain how these complete sets of composable relations can be joined. This is simply a matter of applying the appropriate mathematical operation to the provided terms.

$$\begin{aligned} R(v, C^{+}) &= \{v = C^{+}(f_{1+}), \ldots, v = C^{+}(f_{m+})\} \wedge \\ R(v, C^{-}) &= \{v = C^{-}(f_{1-}), \ldots, v = C^{-}(f_{n-})\} \rightarrow \\ v &= f_{1+} + \ldots + f_{m+} - (f_{1-} + \ldots + f_{n-}) \end{aligned} \tag{9}$$

$$\begin{aligned} R(v, C^{\times}) &= \{v = C^{\times}(f_{1\times}), \ldots, v = C^{\times}(f_{m\times})\} \wedge \\ R(v, C^{\div}) &= \{v = C^{\div}(f_{1\div}), \ldots, v = C^{\div}(f_{n\div})\} \rightarrow \\ v &= \frac{1 \times f_{1\times} \times \ldots \times f_{m\times}}{f_{1\div} \times \ldots \times f_{n\div}} \end{aligned} \tag{10}$$

$$\begin{aligned} R(v, C^{\text{if}}) =&\{v = C^{\text{if},p_1}(a_1, f_1), \ldots, v = C^{\text{if},p_m}(a_m, f_m)\} \wedge \\ R(v, C^{\text{else}}) =&\{v = C^{\text{else}}(f_{\text{else}})\} \wedge p_1 > \ldots > p_m \rightarrow \\ v =&\text{if } a_1 \text{ then } f_1, \text{ else } \ldots, \text{if } a_m \text{ then } f_m, \text{else } f_{\text{else}} \end{aligned} \tag{11}$$

**Property definitions** *Property definitions* describe features of interest to the application requiring a scenario model. A property definition $\Pi$ is a tuple $\langle P^s, \Phi, \pi \rangle$ where $P^s = \{p^s_1, \ldots p^s_m\}$ is a set of *source-participants*, a predicate calculus sentence $\Phi$ whose free variables are elements of $P^s$, and $\pi$ is a relation, whose free variables are also elements of $P^s$, such that

$$\forall p^s_1, \ldots, \forall p^s_m \Phi \rightarrow \pi$$

A typical example of a feature of interest is the requirement that a certain variable in the model is endogenous or exogenous. To be more specific, the property definitions below describe when a variable `?v` is *endogenous* and *exogenous* respectively.

```
(defproperty endogenous
  :source-participants ((?v :type variable))
  :structural-condition ((or (== ?v *) (d/dt ?v *)))
  :property (endogenous ?v))
```

```
(defproperty exogenous
  :source-participants ((?v :type variable))
  :structural-condition ((not (endogenous ?v)))
  :property (exogenous ?v))
```

The first definition states that whenever either `?v = *` or $\frac{d}{dt}$`?v = *` is true (where `*` matches any constant or formula), `?v` is deemed to be endogenous. The second property definition indicates that a variable is said to be *exogenous* if such an object exists and it is not endogenous.

By describing such features formally in the knowledge base, property definitions enable them to be imposed as criteria on the selection of scenario models. In this way, the variable describing the size of a particular population in an eco-system, for instance, can be forced to be endogenous.

Note that required properties can be specified in two different ways: either *globally* as goals for the scenario model construction or locally as a *required purpose* of a certain model fragment. The latter use of model properties will be illustrated later.

**Model fragments** Model fragments are the building blocks with which scenario models are constructed. A model fragment $\mu$ is a tuple $\langle P^s, P^t, \Phi^s, \Phi^t, A, \Pi \rangle$ where $P^s = \{p^s_1, \ldots p^s_m\}$ is a set of variables called *source-participants*, $P^t = \{p^t_1, \ldots, p^t_n\}$ is a set of variables called *target-participants*, $\Phi^s = \{\phi^s_1, \ldots, \phi^s_v\}$ is a set of relations, called *structural conditions*, whose free variables are elements of $P^s$, $\Phi^t = \{\phi^t_1, \ldots, \phi^t_x\}$ is a set of relations, called *postconditions*, whose free variables are elements of $P^s \cup P^t$, $A = \{a_1, \ldots, a_y\}$ is a set of relations, called *assumptions*, and $\Pi =$ is a set of relations, called *purpose-required properties*, such that:

$$\forall \phi^t_i \in \Phi^t, \forall p^s_1, \ldots, \forall p^s_m, \exists p^t_1, \ldots, \exists p^t_n,\ \phi^s_1 \wedge \ldots \wedge \phi^s_v \rightarrow (a_1 \wedge \ldots \wedge a_y \rightarrow \phi^t_i) \qquad (12)$$

$$\forall \pi \in \Pi, \forall p^s_1, \ldots, \forall p^s_m, \forall p^t_1, \ldots, \forall p^t_n,\ \phi^s_1 \wedge \ldots \wedge \phi^s_v \wedge a_1 \wedge \ldots \wedge a_x \wedge \neg\pi \rightarrow \bot \qquad (13)$$

Note that, in this work, each property definition $\langle P^s, \Phi, \pi \rangle$ is equivalent to a model fragment $\langle P^s, \{\}, \Phi, \{\pi\}, \{\}, \{\} \rangle$.

For example, the model fragment below states that a population `?p` can be described by two variables `?p-size` (describing the size of `?p`) and `?p-change` (describing the rate of change in population size) and a differential equation

$$\frac{d}{dt}\texttt{?p-size} = \texttt{?p-change}$$

The usage of this partial scenario model is subject to two conditions: (1) the growth phenomenon is relevant with regard to `?p`, and (2) the variable `?p-change` is endogenous in the eventual scenario model. The former requirement is indicated by the relevance assumption and the latter by the purpose-required property:

```
(defModelFragment population-growth
  :source-participants ((?p :type population))
  :assumptions ((relevant growth ?p))
  :target-participants ((?p-size :type variable)
                        (?p-change :type variable))
  :postconditions ((size-of ?p-size ?p)
                   (change-of ?p-change ?p)
                   (d/dt ?p-size ?p-change))
  :purpose-required ((endogenous ?p-change)))
```

The purpose-required property is usually satisfied by additional model fragments, such as the one below:

```
(defModelFragment logistic-population-growth
  :source-participants ((?p :type population)
                        (?p-size :type variable)
                        (?p-change :type variable))
  :structural-conditions ((size-of ?p-size ?p)
                          (change-of ?p-births ?p))
  :assumptions ((model ?p-size logistic))
  :target-participants ((?r :type parameter)
                        (?k :type variable)
                        (?d :type variable))
  :postconditions ((capacity-of ?k ?p)
                   (density-of ?d ?p-size)
                   (== ?d (C-add (/ ?p-size ?k)))
    (== ?p-change (- (* ?r ?p-size (- 1 ?d))))))
```

Model fragments are rules of inference that describe how new knowledge can be derived from existing knowledge by committing the emerging model to certain assumptions. They are used to generate a space of possible models. Model fragments are *instantiated* by matching source-participants to existing participants in the scenario or an emerging model, and by matching the structural conditions to corresponding relations. For each possible instantiation, a new instance is generated for each of the target-participants, and where necessary, new instances are also created for the postconditions and assumptions. Such instances, as well as the inferential relationships between the instances of the source-participants, structural conditions and assumptions on the one hand, and those of the target-participants and postconditions on the other, are stored in an ATMS, forming the *model space*. This is to be further explained in section 3.3.1.

A model fragment is said to be *applied* if it is instantiated and the underlying assumptions hold. If a model fragment is applied, the instances of the target-participants and postconditions corresponding to the instantiation of that model fragment must be added to the resulting model. With respect to the above example, the model fragment that implements the logistic population growth model is instantiated whenever variables exist that describe the size and change in a population, and it is applied if the logistic model for population size has also been selected.

Note that in most compositional modellers, such as the ones devised by Heller and Struss (1998, 2001); Levy, Iwasaki and Fikes (1997); Nayak and Joskowicz (1996); and Rickel and Porter (1997), model fragments represent direct translations of components of physical systems into influences between variables. Because the compositional modeller presented herein aims to serve as an ecological model repository, the contents of the model fragments employed differs from that of conventional compositional modellers in two important regards:

Firstly, model fragments contain partial models describing certain phenomena instead of influences. These partial models normally correspond to those developed in ecological modelling research. Typical examples include the logistic population growth model (Verhulst, 1838) and the Holling predation model (Holling, 1959) devised in the population dynamics literature.

Secondly, the partial models contained in the model fragments often need to be composed incrementally. For example, the aforementioned sample model fragment `logistic-population-growth` requires an emerging scenario model, which may be generated by the other sample model fragment `population-growth`. Thus, one model fragment, e.g. `logistic-population-growth`, can expand on the partial model contained in another, e.g. `population-growth`. Because of this feature, it is (correctly) presumed that no model fragment $\mu$ generates new relations that are preconditions of model fragments that $\mu$ expands on. Violating this presumption would make little sense in the context of the present application as it would imply a recursive extension of an emerging scenario model with the same set of variables and equations.

### 3.2.4 PARTICIPANT CLASS DECLARATION AND PARTICIPANT TYPE HIERARCHIES

In general, participant classes need not be defined. However, certain types of participant may be described in terms of other interesting participants, irrespective of the modelling choices. This feature provides syntactic sugar for describing important relations between participants, making it easier to declare required properties of a scenario model in terms of the participants of the scenario. For example, the behaviour of a population may be described in terms of population size and growth rate variables:

```
(defEntity population
  :participants (size growth-rate))
```

Participant class declarations may also be employed within model fragments to provide a more specific definition of the meaning of the source-participants and the target-participants. In this way, participant specifications are constrained to be a feature of another participant by means of the `:entity` statement, as the following example illustrates:

```
(defModelFragment define-population-growth-phenomenon
  :source-participants ((?p :type population))
  :target-participants
    ((?ps :type stock :entity (size ?p))
     (?pg :type variable :entity (growth-rate ?p))
     (?pb :type flow)
     (?pd :type flow))
  :assumptions ((relevant growth ?p))
  :postconditions ((== ?pg (- ?pb ?pd))
                   (flow ?pb source ?pl)
                   (flow ?pd ?pl sink)))
```

Furthermore, participant class declarations may define one class to be an immediate subclass of another. For example, the population participant class of holometabolous insects (e.g. butterflies) may be defined as a subclass of the population participant class:

```
(defEntity holometabolous-insect-population
  :subclass-of (population)
  :participants
    (larva-number pupa-number adult-number))
```

In this way, a participant type hierarchy is defined. Each subclass inherits all participants of its superclasses (i.e. its immediate superclass and superclasses of superclasses).

In summary, a *participant class declaration* is a tuple $\Pi = \langle \Pi_S, P \rangle$ where $\Pi_S$ is a participant class, called the immediate superclass of the participant class and $P$ is a set of participants classes that describe important features of the participant class.

### 3.3 Inference

The compositional modelling method presented herein employs a four step inference procedure:

1. *Model space construction*. The model space is an ATMS that efficiently stores all the participants, relations and model design decisions (represented in the form of relevance and model assumptions) that may be part of the final scenario model, as well as the conditions under which each of these participants and relations must or must not be part of the scenario model.

2. *aDCSP construction*. The model space contains a number of hard constraints on the participants and relations that may be combined. This inference step extracts such restrictions and translates them into an aDCSP.

3. *Inclusion of order-of-magnitude preferences*. Preferences are associated with relevance and model assumptions in the scenario space as they reflect the relative appropriateness of these assumptions, resulting in an aDPCSP.

4. *Scenario model selection*. This inference step solves the aDPCSP. The resulting solutions correspond to scenario models that are consistent according to the domain knowledge and optimise the overall preference with respect to the order-of-magnitude preference calculus.

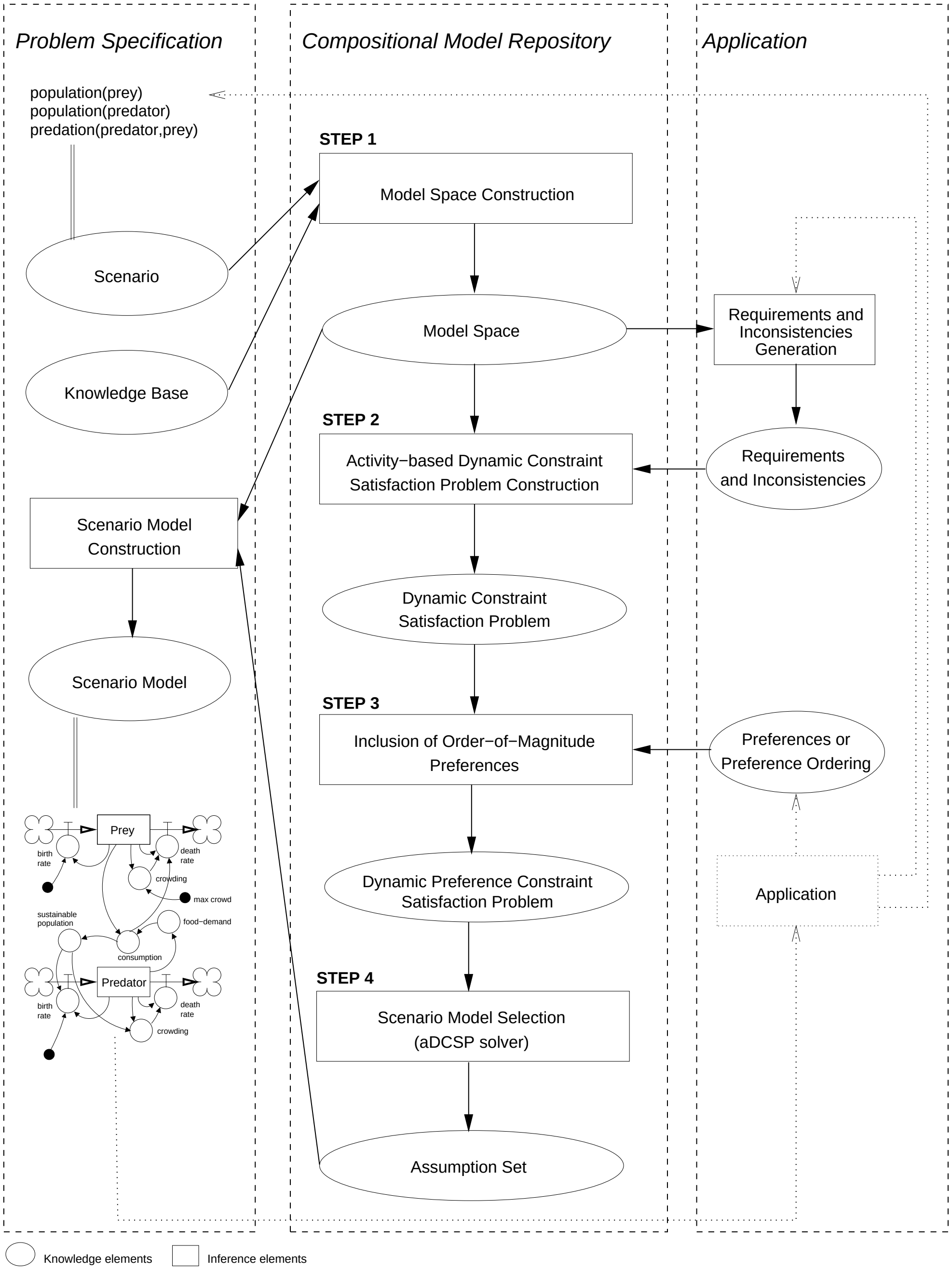

Figure 4: Inference procedures of the compositional modeller

These four steps correspond to the four squares of the compositional model repository in Figure 4

In this section, each of these inference steps is discussed in detail and illustrated by means of simple examples. The next section contains a more detailed example and shows how this procedure can be applied to a non-trivial ecological modelling domain.

#### 3.3.1 SCENARIO + KNOWLEDGE BASE = MODEL SPACE

As previously stated, the aim of a compositional modeller is to translate a scenario into a scenario model. Both are representations of the system of interest though they model the system at a different level of detail. The knowledge base provides the foundation for translation. All the scenario models that can be constructed from the given scenario, with regard to the knowledge base, are stored in the model space.

A model space is an ATMS (de Kleer, 1986) containing all the participants, relations and assumptions that can be instantiated from a given scenario. In this work, the generalised version of the ATMS, as introduced by de Kleer (1988), is employed as it allows the use of negations of nodes in the justifications. The algorithm GENERATEMODELSPACE($\langle O, R \rangle$) describes how such a model space can be created from a scenario $\langle O, R \rangle$. It first initialises the model space $\theta$ with the participant instances ($O$) and the relation instances ($R$) from the scenario. Then, for each model fragment whose source-participants and structural conditions match participants and relations already in $\theta$, new instances of its target-participants, assumptions and postconditions are added to $\theta$. Because each property definition $\langle P^s, \Phi, \pi \rangle$ is equivalent to a model fragment $\langle P^s, \{\}, \Phi, \{\pi\}, \{\}, \{\} \rangle$, this procedure applies to property definitions as well as model fragments. Matching the source-participants and structural conditions of a model fragment $\mu$ to the emerging model space is performed by the function match$(\mu, \theta, \sigma)$ as specified below, where $\mu$ is the model fragment being matched, and $\sigma$ is a substitution from the source-participants of $\mu$ to participant instances.

$$\text{match}(\mu, \theta, \sigma) = \begin{cases} \text{true} & \text{if} \quad \sigma = \{p_1^s/o_1, \ldots, p_m^s/o_m\} \wedge \\ & \qquad P^s(\mu) = \{p_1^s, \ldots, p_m^s\} \wedge \\ & \qquad o_1 \in \theta \wedge \ldots \wedge o_m \in \theta \wedge \\ & \qquad \forall \phi \in \Phi^s(\mu), \sigma\phi \in \theta \\ \text{false} & \text{otherwise} \end{cases}$$

Each match, specified by a model fragment $\mu$ and a substitution $\sigma$, is processed as follows:

- For each assumption $a \in A(\mu)$, a new node, denoting the assumption instance $\sigma a$, is created and added to $\theta$.
- Then, a new node $n_{(\sigma,\mu)}$, denoting the instantiation of $\mu$ via substitution $\sigma$, is created, added to $\theta$ and justified by the implication:

$$(\wedge_{a \in A(\mu)} \sigma a) \wedge (\wedge_{p \in P^s(\mu)} \sigma p) \wedge (\wedge_{\phi \in \Phi^s(\mu)} \sigma \phi) \rightarrow n_{(\sigma,\mu)}$$

- Finally, a new instance for each target-participant $p \in P^t(\mu)$ and for each postcondition $\phi \in \Phi^t(\mu)$, provided $\sigma\phi$ does not already exist in the model space $\theta$, is created. For the target-participants, this involves creating a new symbol for each new participant instance with the function gensym() and extending $\sigma$ with the substitution $\{p/\text{gensym}()\}$. A new node $n$

**Algorithm 1:** GENERATEMODELSPACE($\langle O, R \rangle$)

$\theta \leftarrow$ new ATMS;
**for each** $o \in O$, add-node($\theta, o$);
**for each** $r \in R$, add-node($\theta, r$);
**for each** $\mu, \sigma$, match($\mu, \theta, \sigma$)
**do**
- justification $\leftarrow \emptyset$;
- **for each** $a \in A(\mu)$
  - **do**
    - newnode $\leftarrow$ add-node($\theta, (\sigma a)$);
    - justification $\leftarrow$ justification $\cup$ {newnode};
- **for each** $p \in P^s(\mu)$
  - **do** justification $\leftarrow$ justification $\cup$ {find-node($\theta, (\sigma p)$)};
- **for each** $\phi \in \Phi^s(\mu)$
  - **do** justification $\leftarrow$ justification $\cup$ {find-node($\theta, (\sigma \phi)$)};
- add-node($\theta, n_{(\sigma,\mu)}$);
- add-justification($\theta, n_{(\sigma,\mu)}, \wedge_{n \in \text{justification}} n$);
- **for each** $p \in P^t(\mu)$
  - **do**
    - $\sigma \leftarrow \sigma \cup \{p/\text{gensym}()\}$;
    - $o \leftarrow$ add-node($\theta, (\sigma p)$);
    - add-justification($\theta, o, n_{(\sigma,\mu)}$);
- **for each** $\phi \in \Phi^t(\mu)$
  - **do**
    - **if** ($\sigma \phi \in \theta$)
      - **then** $o \leftarrow$ get-node($\theta, (\sigma \phi)$);
      - **else** $o \leftarrow$ add-node($\theta, (\sigma \phi)$);
    - add-justification($\theta, o, n_{(\sigma,\mu)}$);

**for each** $n_1, \ldots, n_m$, inconsistent($\{n_1, \ldots, n_m\}$)
**do** add-justification($\theta, n_\perp, n_1 \wedge \ldots \wedge n_m$);

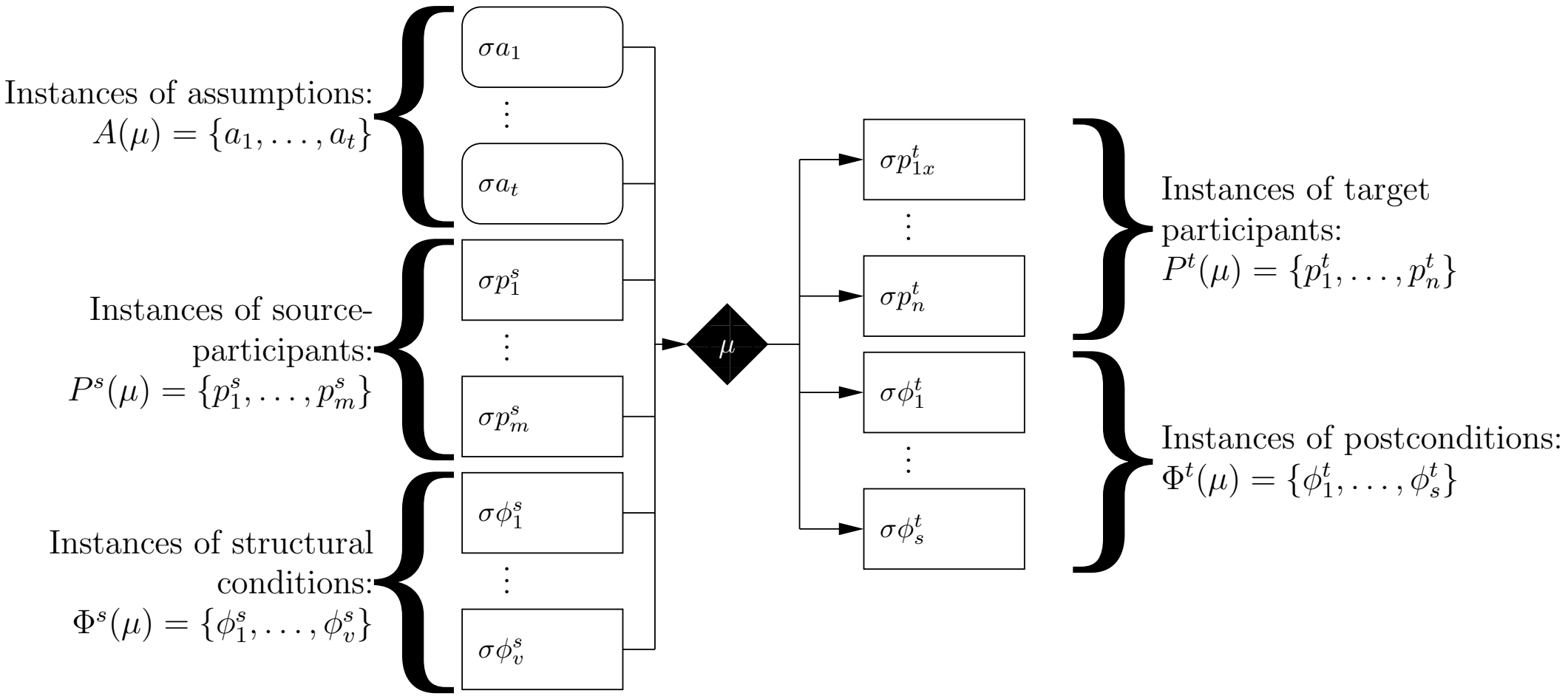

Figure 5: Model fragment application

is created and added to $\theta$ for each new participant instance $\sigma p$ and for each new instantiated relation $\sigma \phi$. Each of these nodes is justified by the implication $n_{(\sigma,\mu)} \rightarrow n$.

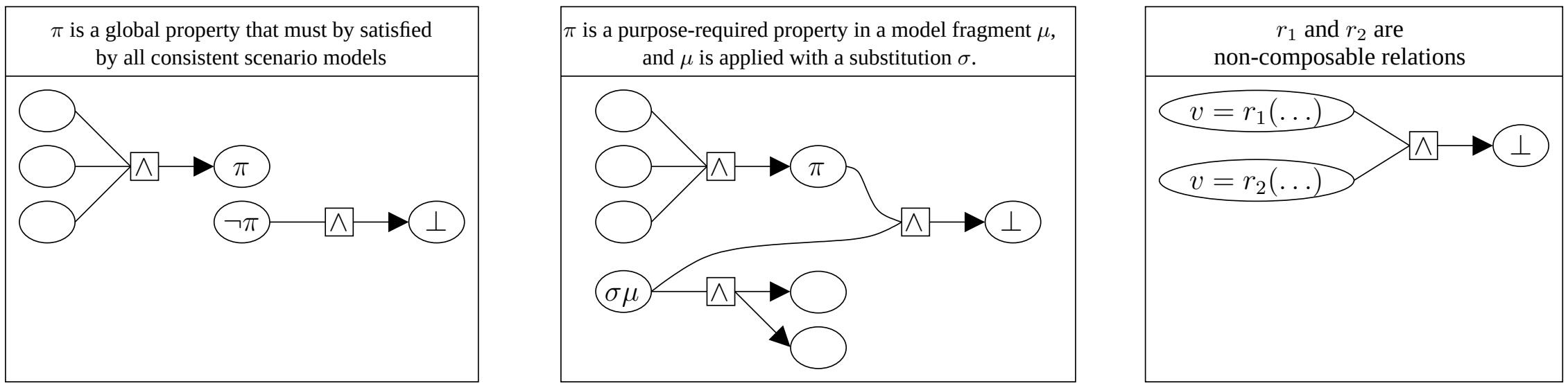

(a) Inconsistency caused by a global property

(b) Inconsistency caused by a purpose-required property

(c) Inconsistency caused by non-composable relations

Figure 6: Sources of inconsistency

To illustrate this procedure, Figure 5 shows a graphical representation of the inferences that are constructed by applying a model fragment $\mu = \langle P^s, P^t, \Phi^s, \Phi^t, A, \{\}\rangle$ with respect to a substitution $\sigma$.

Once all possible applications of model fragments have been exhausted, the inconsistencies in the model space are identified and recorded in the ATMS. In the algorithm, nogoods are generated for each set $\{n_1, \ldots, n_m\}$ of inconsistent nodes, denoted inconsistent$(\{n_1, \ldots, n_m\})$. There are three sources of inconsistencies that are each reported to the ATMS in a different way:

- *Global properties*: Let $\pi$ be an instance of a global property that any scenario model must satisfy. Then, any combination of assumptions and negations of assumptions that prevents $\pi$ from being satisfied is inconsistent. Therefore, inconsistent$(\{\neg\pi\})$ must be reported for any required global property $\pi$. This type of inconsistency is depicted in Figure 6(a).

- *Purpose-required properties*: Any application of a model fragment $\mu$ without satisfying its purpose-required properties $\Pi(\mu)$ yields an inconsistency (see (13)). Hence, for each node $n_{(\sigma,\mu)}$ denoting the instantiation of $\mu$ via substitution $\sigma$, and for each node $n_{\sigma\pi}$ describing the appropriate instance of a purpose-required property $\pi \in \Pi(\mu)$, inconsistent$(\{n_{(\sigma,\mu)}, \neg n_{\sigma\pi}\})$ is reported. This type of inconsistency is depicted in Figure 6(b).

- *Non-composable relations*: In any mathematical formalism designed to describe simulation models of dynamic systems, certain combinations of relations may over-constrain the model, and hence, be unsuitable for generating the behaviour of a system of interest. Within the system dynamics and ODE formalisms used in this paper, assignments of relations to the same variable are only composable if those relations are explicitly deemed composable. In other words, two relations $v = r_i$ and $v = r_j$ can only be combined with one another if $r_i$ and $r_j$ are composable. Examples of pairs of non-composable relations include

  $x = C^+(y)$ and $x = C^\times(z)$ because $C^+$ and $C^\times$ relations are not composable, and
  $a = C^+(b)$ and $a = c + d$ because $c + d$ is not a composable relation.

  Combinations of such non-composable relations must be reported as an inconsistency as well. This type of inconsistency is depicted in Figure 6(c).

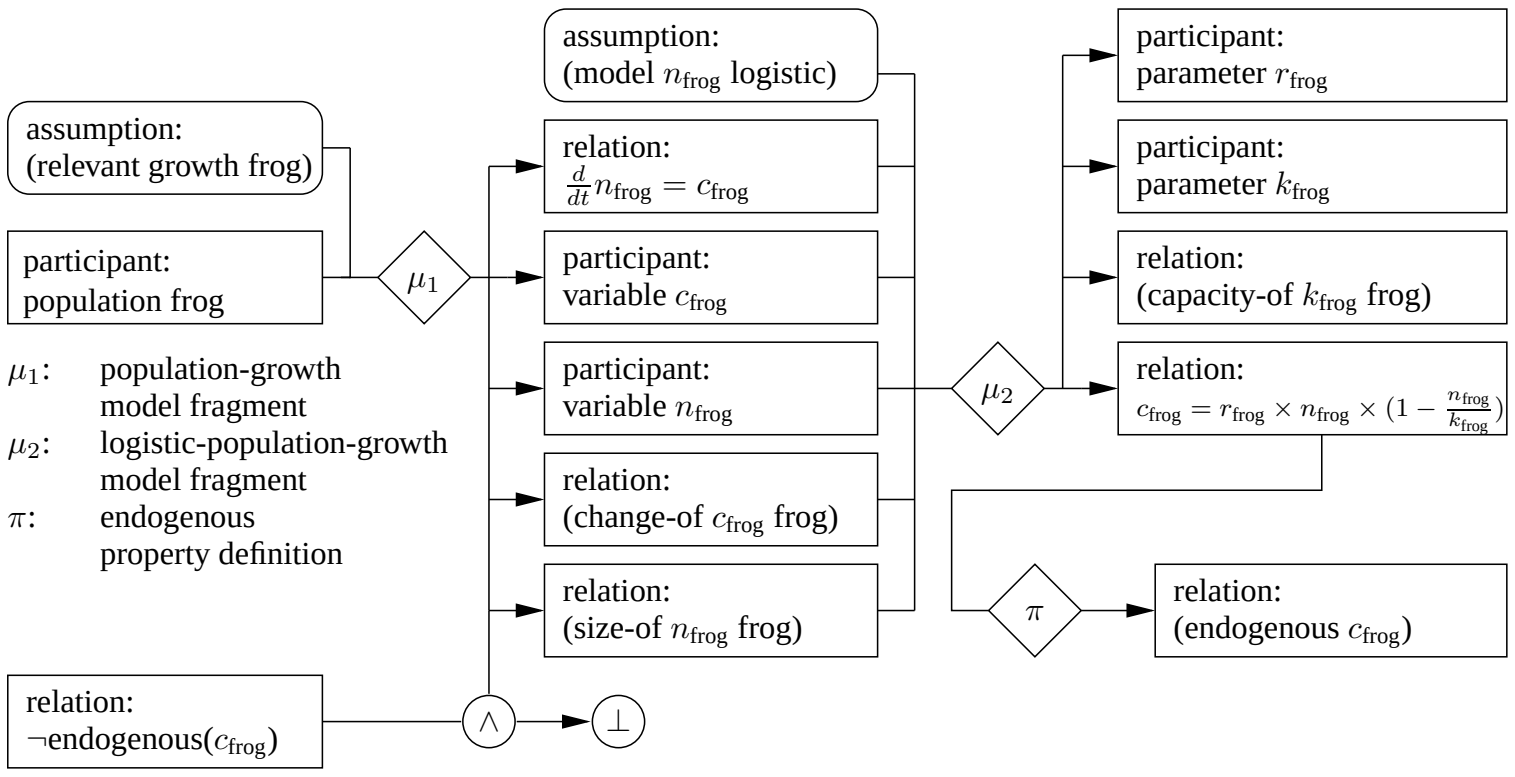

Figure 7: Partial model space

To illustrate the model space construction algorithm, Figure 7 presents a small sample model space. It results from the application of the `population-growth` and `logistic-population-growth` model fragments and the `endogenous` property definition, which were described earlier, for a single population "frog". If a larger scenario involving multiple populations and relations between these populations were specified, a similar partial model space would be generated for each individual population.

### 3.3.2 FROM MODEL SPACE TO aDCSP

Once the model space has been constructed, it can be translated into an aDCSP. The translation procedure, summarised as algorithm CREATEADCSP(), consists of three steps as described below:

**Algorithm 2:** CREATEADCSP()

**comment:** $\sigma$ is the set of substitutions

$\sigma \leftarrow \{\}$;
**comment:** Generate attributes and domains

**for each** $A$, assumption-class$(A)$
**do**
- $x \leftarrow$ create-attribute$()$;
- $D(x) \leftarrow \{\}$;
- $\sigma \leftarrow \sigma \cup \{A/x\}$;
- **for each** $a \in A$
  **do**
  - $v \leftarrow$ create-value$()$;
  - $D(x) \leftarrow D(x) \cup \{v\}$;
  - $\sigma \leftarrow \sigma \cup \{a/x : v\}$;

**comment:** Generate activity constraints

**for each** $A$, assumption-class$(A)$
**do**
- $s \leftarrow$ subject$(A)$;
- **for each** $\{a_{1\top}, \ldots, a_{p\top}, \neg a_{1\bot}, \ldots, \neg a_{q\bot}\} \in \mathcal{L}(s)$
  **do** add$(\sigma a_{1\top} \wedge \ldots \wedge \sigma a_{p\top} \wedge \sigma\neg a_{1\bot} \wedge \ldots \wedge \sigma\neg a_{q\bot} \rightarrow$ active$(\sigma A))$;

**comment:** Generate compatibility constraints

**for each** $\{a_{1\top}, \ldots, a_{p\top}, \neg a_{1\bot}, \ldots, \neg a_{q\bot}\} \in \mathcal{L}(n_\bot)$
**do** add$(\sigma a_{1\top} \wedge \ldots \wedge \sigma a_{p\top} \wedge \sigma\neg a_{1\bot} \wedge \ldots \wedge \sigma\neg a_{q\bot} \rightarrow \bot)$;

1. *Generate the attributes and domain values from the assumptions*. The aDCSP attributes correspond to the underlying assumption classes (i.e. groups of assumptions indicating alternative choices with regards to the same model construction decision). A relevance assumption and its negation jointly form an assumption class. For example, $A_1 = \{$`(relevant growth frog)`, $\neg$`(relevant growth frog)`$\}$ specifies such an assumption class. The set of model assumptions involving the same participants/relations, but with different model names and hence different descriptions, also form an assumption class. For instance, $A_2 = \{$`(model` $\mathtt{n}_{\text{frog}}$ `exponential)`, `(model` $\mathtt{n}_{\text{frog}}$ `logistic)`, `(model` $\mathtt{n}_{\text{frog}}$ `other)`$\}$, where $\mathtt{n}_{\text{frog}}$ is a variable denoting the size of a population, specifies such an assumption class. Running this step of the algorithm, an attribute is created for each assumption class, with the domain of such an attribute consisting of all assumption instances in the assumption class.

2. *Create activity constraints*. The attributes and domain values generated in the previous step are only meaningful in situations where the participant and/or relation instances contained in the arguments of the corresponding assumptions exist. For example, the assumption `(model` $\mathtt{n}_{\mathtt{frog}}$ `logistic)` is only relevant if the participant instance $\mathtt{n}_{\text{frog}}$ exists. Clearly, all assumptions within one assumption class have the same participant and/or relation instances as their arguments. Because each assumption class corresponds to one attribute, the attribute can be activated if and only if the participant and/or relation instances associated with the related assumption class are active. Therefore, this step creates activity constraints that activate an attribute based on the conjunction of the environments contained within the labels of the participants/relations of the assumption class. For instance, as can be deduced from Figure 7, $\mathtt{n}_{\text{frog}}$ is activated when `(relevant growth frog)` is committed. Thus, the attribute corresponding to assumption class $A_2$, defined in step 1, is activated under the attribute value assignment associated with the `(relevant growth frog)` assumption.

3. *Create compatibility constraints*. In the ATMS (or model space), all sources of inconsistencies are contained in the label of the nogood node. Therefore, the compatibility constraints are created directly by translating the environments in the label $\mathcal{L}(\perp)$ into the corresponding conjunctions of attribute-value assignments.

### 3.3.3 aDCSP + PREFERENCES = aDPCSP

The aDCSP produced as above formalises the hard requirements imposed upon the scenario models. Among the scenario models that meet these requirements, some may be better than others, because the underlying model design decisions may be deemed more appropriate by the user. Preferences that express this (relative) level of appropriateness are attached to the assumptions that describe the model design decisions, and by extension, to the attribute-value pairs in the aDCSP. As discussed in section 2, such an extension to the aDCSP constitutes an aDPCSP.

More specifically, it is worth recalling that in section 2.2 an order-of-magnitude preference calculus is presented that enables representation and reasoning with subjective user preferences for different relevance and modelling assumption. Next, section 2.3 introduces a solution algorithm for aDPCSPs that include an aDCSP, such as the ones constructed with the approach of section 3.3.2, and are extended with subjective user preferences for alternative design decisions.

### 3.4 Outline analysis of complexity

The complexity of the work arises from four major sources: 1) model space construction, 2) label propagation in the ATMS, 3) model space to aDCSP translation, and 4) aDPCSP solution.

GENERATEMODELSPACE($\langle O, R\rangle$) essentially performs a fixed sequence of instructions and produces a small set of nodes and inferences for each match of a model fragment. Therefore, its time and space complexity are linear with respect to the number of possible matches of model fragments. CREATEADCSP() extracts certain information from the model space and rewrites it in a different formalism without further manipulations. Therefore, its time and space complexity are linear with respect to the size of the model space.

The label propagation algorithm of an ATMS is known to have an exponential time complexity. However, because the model space is built up incrementally (by GENERATEMODELSPACE($\langle O, R\rangle$)) from the root nodes of the ATMS network (i.e. those that correspond to facts and have no antecedents) to the leaf nodes (i.e. those that have have no consequents, other than the nogood node) and because the inconsistencies are added at the end, this complexity only increases exponentially with the depth of the network and the number of participants and relations in individual model fragments, rather than with the size of the model space. This fact significantly limits the complexity impact of label propagation. Firstly, the depth of the ATMS network is restricted by the domain. In many conventional compositional modellers, where model fragments are direct translations from scenario components to scenario model equations, this depth would be only one. Empirically, constructing the model space for sophisticated eco-systems, the depth of a model space never exceeded 8. Secondly, the size of the individual model fragments does not change significantly with the size of the knowledge base.

The fourth and final source of complexity is driven by the fact that the constraint satisfaction algorithm must determine a consistent combination of assumptions in the model space. The space of attribute value assignments increases exponentially with the size of the number of assumptions and hence, with the model space. Thus, the overall complexity of the present approach is largely dominated by the constraint satisfaction algorithm employed.

If the user does not specify any preference, the CSP is an aDCSP. Recently, a number of efficient methods have been devised for solving aDCSPs as presented by Minton et al. (1992); Mittal and Falkenhainer (1990); and Verfaillie and Schiex (1994). This helps minimise the overhead incurred for compositional modelling.

With preferences, the CSP becomes an aDPCSP. As argued in section 2, this presents a new problem that has not yet been studied in detail. In this work, an A* algorithm has been proposed to implement the CSP solution method. This approach is known to be the most efficient in terms of the proportion of the search space the algorithm needs to explore before finding an optimal solution, when compared to other search methods that are based on the same heuristic (Hart et al., 1968). A disadvantage is that it incurs an exponential space complexity. As explained by Miguel and Shen (2001a, 2001b); and Tsang (1993), a wide range of alternative solution techniques exist for ordinary CSPs and many of these could also be extended to solve aDPCSPs. A detailed examination of these techniques is a topic of future research.

### 3.5 Automated modelling and scientific discovery

As mentioned previously, a compositional model repository is designed in order to compose models from a system's structure and relevant domain knowledge. As such, this approach gives rise to a po-tentially beneficial means to operationalise the outcomes of scientific discovery. More specifically, the resultant compositional model repositories will allow existing knowledge on model construction to be applied to unexperienced scenarios and to support investigation into situations which may be physically difficult to replicate or create but which may be synthesised in computational representations.

The present work has been applied to the vegetation component of the MODMED $n$-species model (Legg, Muetzelfeldt, & Heathfield, 1995). This $n$-species model offers a system dynamics representation of populations of Mediterranean vegetations and of how they are affected by populations of farm animals, climate and environmental management. The purpose of the model is to be instantiated with respect to various Mediterranean communities, and to serve as a component of a very large scale simulation that is designed to simulate the effects of various environmental policies on the Mediterranean landscape. A knowledge base containing approximately 60 model fragments and 4 property definitions has been constructed, on the basis of the most complex parts of the $n$-species model in about two man-weeks. This knowledge base can be employed to reconstruct variations of the $n$-species model to accommodate a variety of possible scenarios, as well as to examine simplifications of the original $n$-species model which exclude certain phenomena.

The compositional model repository is most closely related to the seminal work on compositional modelling (Falkenhainer & Forbus, 1991). That approach has a similar functionality but it is devised specifically for physical systems and relies on a component-connection formalism to represent scenarios.

Another approach which has recently been developed and applied to the ecological domain by Heller and Struss (1998, 2001). This work derives a system's structure from observations of its behaviour and domain knowledge. Therefore, it is able to perform diagnosis of ecological systems and therapy suggestion. Another important distinction of this work from the present study is that it presumes that each process can only be described in just one way instead of allowing multiple alternative models.

In the machine learning community, a number of approaches have been devised by Bradley, Easley and Stolle (2001); Langley et al. (2002); and Todorovski and Džeroski (1997, 2001) to induce sets of differential equations from a) observations of behaviour, b) domain knowledge represented in the form of hypothetical equations, and c) a description of the structure of the system. These approaches aim at scientific discovery by generalising observed behaviour into mathematical models. The specifications of the scenario and the domain knowledge in these methods are similar to those used in this article. This is especially true for the work by Langley et al. (2002); and Todorovski and Džeroski (1997, 2001), because that work has also been applied to population dynamics. However, the internal mechanisms of these approaches are very different as they essentially rely on exhaustive search procedures instead of constraint satisfaction techniques.

## 4. A Population Dynamics Example

The examples used throughout the previous sections were taken from a more extensive application study of the present work. The application was aimed to construct a repository of basic population dynamic models, describing the phenomena of growth, predation and competition. This section presents an overview of how the proposed approach is employed in this application to show the ability of the work to scale to larger problems.

### 4.1 Knowledge base

This subsection illustrates how a set of model fragments can be constructed. The challenge of this task lies in the fact that model fragments must encompass a sufficiently general and reusable component part of the ecological models. In instances of models found in the literature on ecological modelling, the boundaries of the recurring component parts are hidden, and it is therefore up to the knowledge engineer to identify them.

First, a hierarchy of entity types is set up. The system dynamics models shown earlier contain only three types of participant: variables, stocks and flows. Here, stocks and flows are a special type of variable with a predetermined meaning. That is, a flow $f$ into a stock $s$ corresponds to the equation $\frac{d}{dt}s = C^{+}(f)$ and a flow $f$ out of a stock $s$ denotes $\frac{d}{dt}s = C^{-}(f)$. Hence, stocks and flows are defined as subclasses of the participant class variable:

```
(defEntity variable)
(defEntity stock
  :subclass-of (variable))
(defEntity flow
  :subclass-of (variable))
```

The sample properties defined in section 3.2.3, which describe the condition under which a variable is endogenous or exogenous, are employed in this knowledge base:

```
(defproperty endogenous-1
  :source-participants ((?v :type variable))
  :structural-conditions ((== ?v *))
  :property (endogenous ?v))

(defproperty endogenous-2
  :source-participants ((?v :type variable))
  :structural-conditions ((d/dt ?v *))
  :property (endogenous ?v))

(defproperty exogenous
  :source-participants ((?v :type variable))
  :structural-conditions ((not (endogenous ?v)))
  :property (exogenous ?v))
```

The next three model fragments contain the rules of the stock-flow diagrams employed by systems dynamics models. They respectively describe that:

- A flow `?flow` into a stock `?stock` corresponds to the composable differential equation:

$$\frac{d}{dt}\texttt{?stock} = C^{+}(\texttt{?flow})$$

- A flow `?flow` out of a stock `?stock` corresponds to the composable differential equation:

$$\frac{d}{dt}\texttt{?stock} = C^{-}(\texttt{?flow})$$

- A flow `?flow` from one stock `?stock1` to another stock `?stock2` corresponds to the composable differential equations:

$$\frac{d}{dt}\texttt{?stock1} = C^{-}(\texttt{?flow}) \text{ and } \frac{d}{dt}\texttt{?stock2} = C^{+}(\texttt{?flow})$$

```
(defModelFragment inflow
  :source-participants
    ((?stock :type stock)
     (?flow :type flow))
  :structural-conditions
    ((flow ?flow source ?stock))
  :postconditions
    ((d/dt ?stock (C-add ?flow))))

(defModelFragment outflow
  :source-participants
    ((?stock :type stock)
     (?flow :type flow))
  :structural-conditions
    ((flow ?flow ?stock sink))
  :postconditions
    ((d/dt ?stock (C-sub ?flow))))

(defModelFragment inflow
  :source-participants
    ((?stock1 :type stock)
     (?stock2 :type stock)
     (?flow :type flow))
  :structural-conditions
    ((flow ?flow ?stock1 ?stock2))
  :postconditions
    ((d/dt ?stock1 (C-sub ?flow))
     (d/dt ?stock2 (C-add ?flow))))
```

Once the above declarations are in place, the knowledge base of model fragments can be defined. The first model fragment describes the population growth phenomenon. Note that all of the aforementioned growth, predation and competition models contain a stock representing population size and two flows, one flow of births into the stock and another flow of deaths out of the stock. This common feature of models on population dynamics is contained in a single model fragment.

```
(defModelFragment population-growth
  :source-participants
    ((?population :type population))
  :assumptions
    ((relevant growth ?population))
  :target-participants
    ((?size :type stock :name size)
     (?birth-flow :type flow :name births)
     (?death-flow :type flow :name deaths))
  :postconditions
    ((flow ?birth-flow source ?size)
     (flow ?death-flow ?size sink)
     (size-of ?size ?population)
     (births-of ?birth-flow ?population)
     (deaths-of ?death-flow ?population))
  :purpose-required
    ((endogenous ?birth-flow)
     (endogenous ?death-flow)))
```

The variables `?birth-flow` and `?death-flow` become endogenous if the model contains an equation describing birth flow and death flow. These equations differ between population growth models. Two types of population growth model are the exponential growth model (Malthus, 1798), which is shown in Figure 8(a), and the logistic growth model (Verhulst, 1838), which is shown in Figure 8(b). The following two model fragments formally describe these component models:

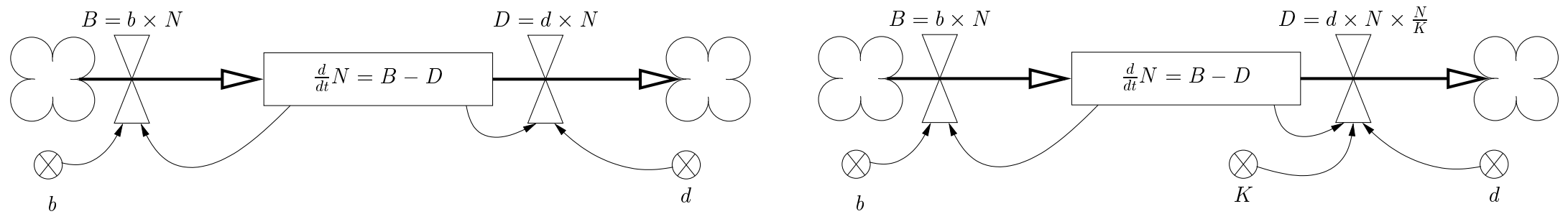

(a) Exponential growth (b) Logistic growth

Figure 8: Population growth models

```
(defModelFragment exponential-population-growth
  :source-participants
    ((?population :type population)
     (?size :type variable)
     (?birth-flow :type variable)
     (?death-flow :type variable))
  :structural-conditions
    ((size-of ?size ?population)
     (births-of ?birth-flow ?population)
     (deaths-of ?death-flow ?population))
  :assumptions
    ((model ?size exponential))
  :target-participants
    ((?birth-rate :type variable :name birth-rate)
     (?death-rate :type variable :name death-rate))
  :postconditions
    ((== ?birth-flow (* ?birth-rate ?size))
     (== ?death-flow (* ?death-rate ?size))))

(defModelFragment logistic-population-growth
  :source-participants
    ((?population :type population)
     (?size :type variable)
     (?birth-flow :type variable)
     (?death-flow :type variable))
  :structural-conditions
    ((size-of ?size ?population)
     (births-of ?birth-flow ?population)
     (deaths-of ?death-flow ?population))
  :assumptions
    ((model ?size logistic))
  :target-participants
    ((?birth-rate :type variable :name birth-rate)
     (?death-rate :type variable :name death-rate)
     (?density :type variable :name total-population)
     (?capacity :type variable :name capacity))
  :postconditions
    ((== ?birth-flow (* ?birth-rate ?size))
     (== ?death-flow (* ?death-rate ?size ?density))
     (== ?density (C-add (/ ?size ?capacity)))
     (density-of ?density ?population)
     (capacity-of ?capacity ?population)))
```

There is one twist in compositional modelling of population growth. Sometimes, the actual growth model is implicitly contained within another type of model. In such cases, the growth phenomenon and the corresponding differential equations are still relevant, but none of the dedicated growth models can be employed. For example, as will be shown later, the Lotka-Volterra predation model comes with its own equations describing growth.

The model fragment `other-growth` allows for an empty growth model, named `other`, to be selected. However, due to the purpose-required property that any instance of `?p-change` must be endogenous, this empty model can only be selected if a growth model is implicitly included elsewhere.

```
(defModelFragment other-growth
  :source-participants
    ((?population :type population)
     (?size :type variable)
     (?birth-flow :type variable)
     (?death-flow :type variable))
  :structural-conditions
    ((size-of ?size ?population)
     (births-of ?birth-flow ?population)
     (deaths-of ?death-flow ?population))
  :assumptions
    ((model ?population other)))
```

In addition to population growth, two other phenomena are included in the knowledge base: predation and competition. Predation and competition relations between species are represented by predicates over the populations: e.g. `(predation foxes rabbits)` and `(competition sheep cows)`. However the existence of a phenomenon does not necessarily mean that it must be contained within the model. It would make little sense to model predation and competition without modelling the size of the populations, because models of these phenomena relate population sizes to one another. Therefore, the incorporation of the predation phenomenon is made dependent upon the existence of variables representing population size. Also, human expert modellers may prefer to leave a phenomenon out of the resulting model. To keep this choice open, the following two model fragments construct a participant representing the phenomena of predation and competition, and make it dependent upon a relevance assumption:

```
(defModelFragment predation-phenomenon
  :source-participants
    ((?predator :type population)
     (?prey :type population)
     (?predator-size :type variable)
     (?prey-size :type variable))
  :structural-conditions
    ((predation ?predator ?prey)
     (size-of ?predator-size ?predator)
     (size-of ?prey-size ?prey))
  :assumptions
    ((relevant predation ?predator ?prey))
  :target-participant
    ((?predation-phenomenon :type phenomenon :name predation-phenomenon))
  :postconditions
    ((predation-phenomenon ?predation-phenomenon ?predator ?prey))
  :purpose-required ((has-model ?predation-phenomenon)))

(defModelFragment competition-phenomenon
  :source-participants
    ((?population1 :type population)
     (?population2 :type population)
     (?size1 :type variable)
     (?size2 :type variable))
  :structural-conditions
    ((competition ?population1 ?population2)
     (size-of ?size1 ?population1)
     (size-of ?size2 ?population2))
```

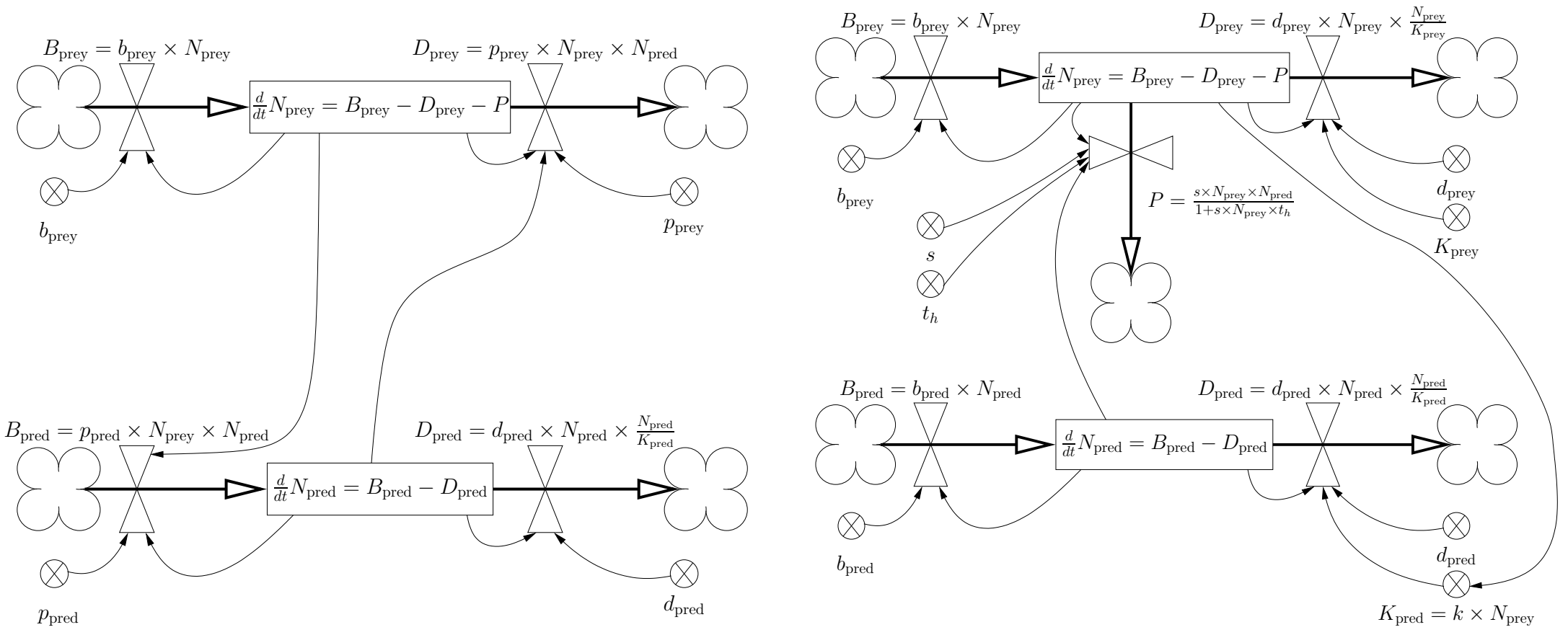

(a) Lotka-Volterra predation (b) Holling predation

Figure 9: Predation models

```
  :assumptions
    ((relevant competition ?population1 ?population2))
  :target-participant
    ((?competition-phenomenon :type phenomenon :name competition-phenomenon))
  :postconditions
    ((competition-phenomenon ?competition-phenomenon ?population1 ?population2))
  :purpose-required
    ((has-model ?competition-phenomenon)))
```

Both model fragments have a purpose-required property of the form `(has-model ?phen)`. This property expresses the condition that a model must exist with respect to a phenomenon:

```
(defproperty has-model
  :source-participants ((?p :type phenomenon))
  :structural-conditions ((is-model-of ?p *))
  :property (has-model ?p))
```

The next two model fragments implement such models (thereby satisfying the above `has-model` purpose-required property) for the predation phenomenon between two populations. They describe two well-known predation models: the Lotka-Volterra model (1925, 1926), which is shown in Figure 9(a), and the Holling model (1959), which is shown graphically in Figure 9(b).

```
(defModelFragment Lotka-Volterra
  :source-participants
    ((?predation-phenomenon :type phenomenon)
     (?predator :type population)
     (?predator-size :type stock)
     (?predator-birth-flow :type flow)
     (?predator-death-flow :type flow)
     (?prey :type population)
     (?prey-size :type stock)
     (?prey-birth-flow :type flow)
     (?prey-death-flow :type flow))
  :structural-conditions
    ((predation-phenomenon ?predation-phenomenon ?predator ?prey)
```

```
     (size-of ?predator-size ?predator)
     (births-of ?predator-birth-flow ?predator)
     (deaths-of ?predator-death-flow ?predator)
     (size-of ?prey-size ?prey)
     (births-of ?prey-birth-flow ?prey)
     (deaths-of ?prey-death-flow ?prey))
  :assumptions
    ((model ?predation-phenomenon lotka-volterra))
  :target-participants
    ((?prey-birth-rate :type variable :name birth-rate)
     (?predator-factor :type variable :name predator-factor)
     (?prey-factor :type variable :name prey-factor)
     (?predator-death-rate :type variable :name death-rate))
  :postconditions
    ((== ?prey-birth-flow (* ?prey-birth-rate ?prey-size))
     (== ?predator-birth-flow (* ?predator-factor ?prey-size ?predator-size))
     (== ?prey-death-flow (* ?prey-factor ?prey-size ?predator-size))
     (== ?predator-death-flow (* ?predator-death-rate ?predator-size))
     (is-model-of lotka-volterra ?predation-phenomenon)))
```

As mentioned earlier, the Lotka-Volterra model introduces its own growth model for the prey and predator populations by assigning specific equations to the variables, which describe changes in the sizes of the predator and prey populations, `?pred-change` and `?prey-change` respectively. Thus, it satisfies the purpose-required property in the application of the `population-growth` model fragment for the `?prey` and `?pred` populations.

```
(defModelFragment Holling
  :source-participants
    ((?predation-phenomenon :type phenomenon)
     (?predator :type population)
     (?predator-size :type stock)
     (?capacity :type variable)
     (?prey :type population)
     (?prey-size :type stock))
  :structural-conditions
    ((predation-phenomenon ?predation-phenomenon ?predator ?prey)
     (size-of ?predator-size ?predator)
     (size-of ?prey-size ?prey)
     (capacity-of ?capacity ?predator))
  :assumptions
    ((model ?predation-phenomenon holling))
  :target-participants
    ((?search-rate :type variable :name search-rate)
     (?handling-time :type variable :name handling-time)
     (?prey-requirement :type variable :name prey-requirement)
     (?predation :type flow :name predation))
  :postconditions
    ((flow ?predation ?prey-size sink)
     (== ?predation
         (/ (* ?search-rate ?prey-size ?predator-size)
            (+ 1 (* ?search-rate ?prey-size ?handling-time))))
     (== ?capacity (C-add (* ?prey-requirement ?prey)))
     (is-model-of holling ?predation-phenomenon)))
```

The Holling model employs a variable denoting the capacity of a population. Such a variable may be introduced by a logistic growth model. In practice, logistic growth models and Holling predation models are often used in conjunction. The compositional modeller need not be aware of such combinations of models, however. All it needs to know is the prerequisites of the individual component models contained within each model fragment.

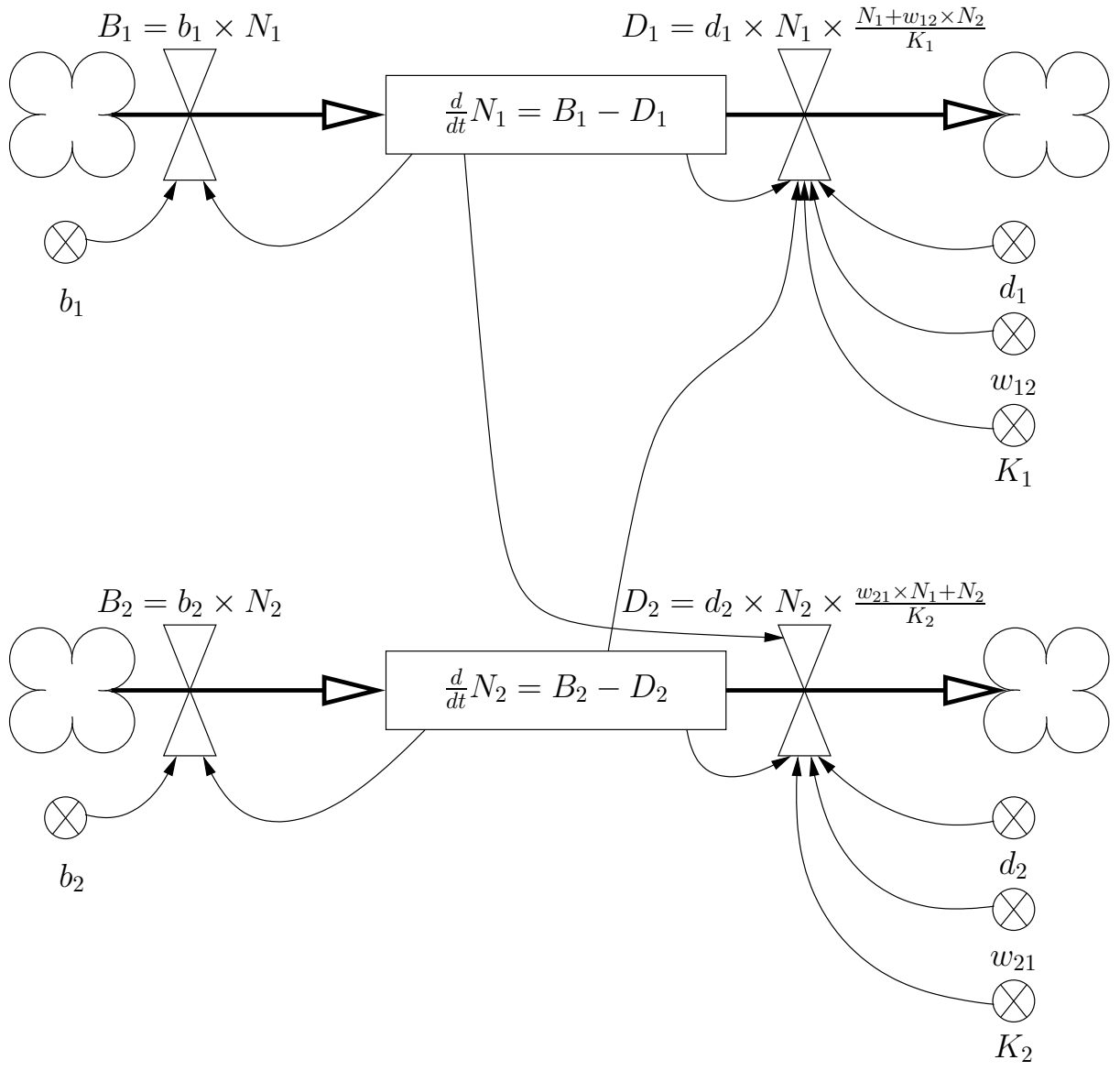

Figure 10: A species competition model

The final model fragment in the knowledge base implements a model of competition between two species. It formally describes the competition model type depicted in Figure 10. As this model fragment contains the only population competition model in the knowledge base, it does not contain a model assumption to represent the model.

```
(defModelFragment competition
  :source-participants
    ((?competition-phenomenon :type phenomenon)
     (?population-1 :type population)
     (?size-1 :type stock)
     (?density-1 :type variable)
     (?capacity-1 :type variable)
     (?population-2 :type population)
     (?size-2 :type stock)
     (?density-2 :type variable)
     (?capacity-2 :type variable))
  :structural-conditions
    ((competition-phenomenon ?competition-phenomenon ?population-1 ?population-2)
     (density-of ?density-1 ?size-1)
     (capacity-of ?capacity-1 ?size-1)
     (density-of ?density-2 ?size-2)
     (capacity-of ?capacity-2 ?size-2))
  :assumptions
    ((relevant competition ?population-1 ?population-2))
  :target-participants
    ((?weight-12 :type variable :name weight)
     (?weight-21 :type variable :name weight))
  :postconditions
    ((== ?density-1 (C-add (/ (* ?weight-12 ?size-2) ?capacity-1)))
     (== ?density-2 (C-add (/ (* ?weight-21 ?size-1) ?capacity-2)))))
```

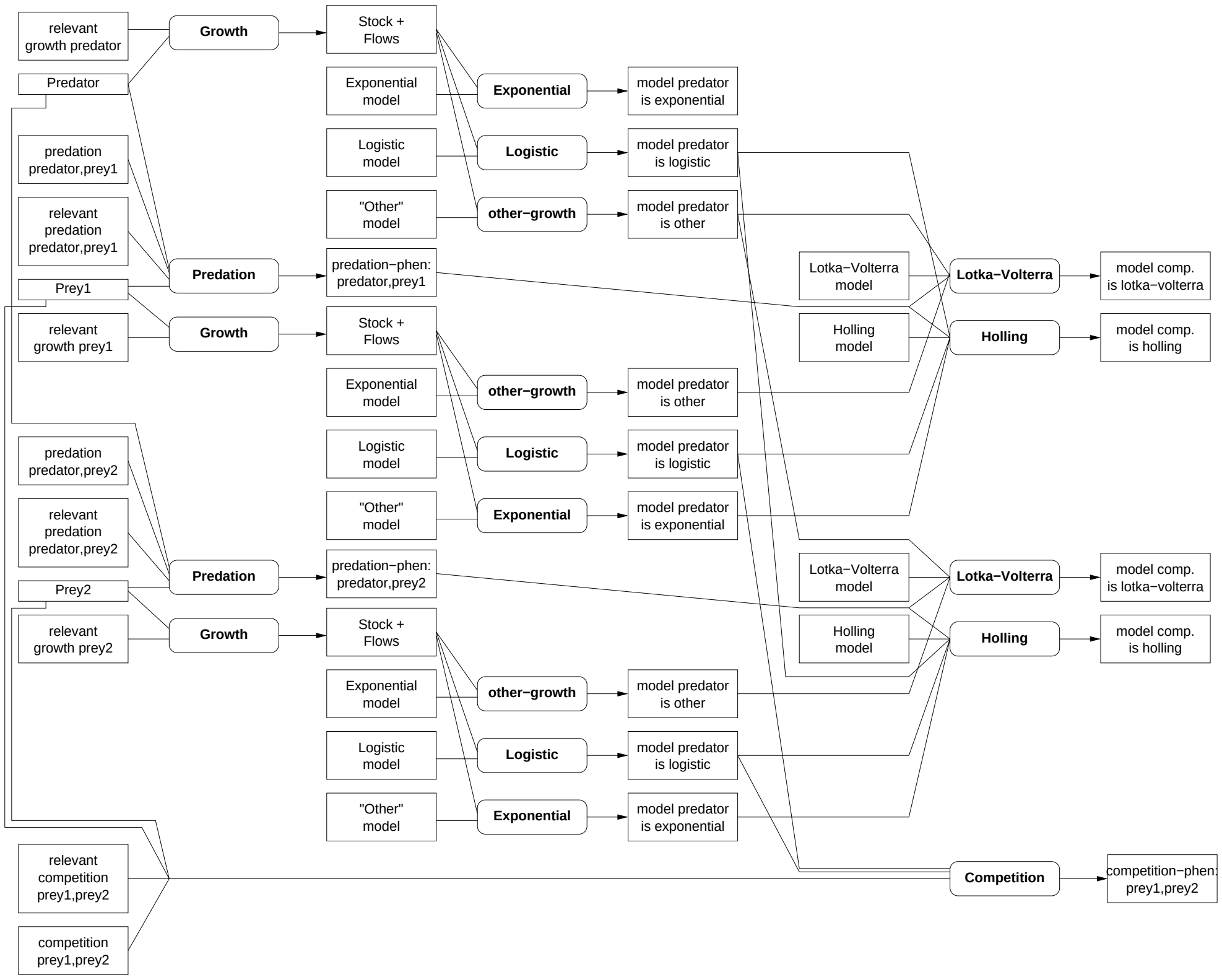

Figure 11: Model space for the 1 predator and 2 competing prey scenario

## 4.2 Model space

A model space is constructed when the knowledge base is instantiated with respect to a given scenario. Consider for example the following scenario, which describes a `predator` population that preys on two other populations, `prey1` and `prey2`, whilst the two prey populations compete with one another:

```
(defScenario pred-prey-prey-scenario
  :entities ((predator :type population)
             (prey1 :type population)
             (prey2 :type population))
  :relations ((predation predator prey1)
              (predation predator prey2)
              (competition prey1 prey2)))
```

The full specification of the model space is too unwieldy to present here but an abstract graphical representation of the model space for this scenario is shown in Figure 11. This model space contains the following knowledge:

- From each of the three populations in the scenario, a set of three population growth models (i.e. `exponential`, `logistic` and `other`) is derived. This inference is dependent upon a relevance assumption of the population growth phenomenon, and a model assumption that corresponds to one of the three population growth models.

- From both predation relations (i.e. `(predation predator prey1)` and `(predation predator prey2)`), and the populations related by them, a set of two predation models (i.e. `Lotka-Volterra` and `Holling`) is derived. This inference is dependent upon a relevance assumption of the predation phenomenon and a model assumption that corresponds to one of the two predation models.

- From the competition relation `(competition prey1 prey2)`, and the populations related by it, a competition model is derived. Because there is only one competition model, the inference of the competition model is only dependent upon a relevance assumption that corresponds to the competition phenomenon.

In addition to the hypergraph of Figure 11, the model space also contains a number of constraints on the conjunctions of assumptions that are consistent. As explained earlier, these stem from two sources: 1) non-composable relations and 2) purpose-required properties. An example will be given of each type.

Let `predation-phen-1` be the predation phenomenon between `predator` and `prey1`, and `prey1-size` be the variable representing the size of the `prey1` population. In this example, the model fragments `exponential-population-growth` and `Lotka-Volterra` will each generate an equation for computing the value of a variable representing the change in `prey1-size`. Because both equations can not be composed, the following inconsistency is generated:

(relevant growth prey1) ∧ (model prey1-size exponential) ∧
(relevant growth predator) ∧ (relevant predation predator prey1) ∧
(model predation-phen-1 lotka-volterra) → ⊥

Inconsistencies also arise from purpose-required properties. For example, if the model fragment `predation-phenomenon` is applicable and the predation relation is deemed relevant, then the purpose-required property `(has-model ?pred-phen)` will become a condition for consistency. Under certain combinations of assumptions, this property may not be satisfied. Say, when the Holling predation and exponential growth models are both selected, the Holling model is not generated because there is no `?capacity` for which `(capacity ?capacity ?pred)` is true. No predation model is created in this case (because the `Holling` model fragment can not be instantiated), even though the predation phenomenon is deemed relevant under this set of assumptions. This is inconsistent with the `has-model` purpose-required property in the `predation-phenomenon` model fragment, and the responsible combination of assumptions is therefore marked as nogood.

(relevant growth predator) ∧ (model predator-size exponential) ∧
(relevant growth prey1) ∧ (model prey1-size exponential) ∧
(relevant predation predator prey1) ∧ (model predation-phen-1 holling) → ⊥

### 4.3 aDPCSP and solution

The resultant model space is translated into an aDCSP to enable the selection of a consistent set of assumptions, using advanced CSP solution techniques. The aDCSP derived from the above model space is depicted in Figure 12.

| Attribute | Meaning |
|---|---|
| $x_1$ | `(relevant growth prey1)` |
| $x_2$ | `(relevant growth prey2)` |
| $x_3$ | `(relevant growth predator)` |
| $x_4$ | `(relevant predation predator prey1)` |
| $x_5$ | `(relevant predation predator prey2)` |
| $x_6$ | `(relevant competition prey1 prey2)` |
| $x_7$ | `(model size-1 *)` |
| $x_8$ | `(model size-2 *)` |
| $x_9$ | `(model size-3 *)` |
| $x_{10}$ | `(model predation-phen-1 *)` |
| $x_{11}$ | `(model predation-phen-2 *)` |

Table 4: Attribute list

| Domain | Content | Meaning |
|---|---|---|
| $D_1$ | $\{d_{1,y}, d_{1,n}\}$ | {population,none} |
| $D_2$ | $\{d_{2,y}, d_{2,n}\}$ | {population,none} |
| $D_3$ | $\{d_{3,y}, d_{3,n}\}$ | {population,none} |
| $D_4$ | $\{d_{4,y}, d_{4,n}\}$ | {(population,population),none} |
| $D_5$ | $\{d_{5,y}, d_{5,n}\}$ | {(population,population),none} |
| $D_6$ | $\{d_{6,y}, d_{6,n}\}$ | {(population,population),none} |
| $D_7$ | $\{d_{7,l}, d_{7,e}, d_{7,o}\}$ | {logistic,exponential,other} |
| $D_8$ | $\{d_{8,l}, d_{8,e}, d_{8,o}\}$ | {logistic,exponential,other} |
| $D_9$ | $\{d_{9,l}, d_{9,e}, d_{9,o}\}$ | {logistic,exponential,other} |
| $D_{10}$ | $\{d_{10,h}, d_{10,lv}\}$ | {Holling,Lotka-Volterra} |
| $D_{11}$ | $\{d_{11,h}, d_{11,lv}\}$ | {Holling,Lotka-Volterra} |

Table 5: The aDCSP for the 1 predator and 2 competing prey scenario: domains and their contents and meaning

This aDCSP contains 11 attributes. They are listed with the corresponding assumption classes in table 4. The first 6 attributes correspond to the notion of relevance phenomenon: 3 population growth phenomena, 2 predation phenomena and 1 competition phenomenon to be precise. The other 5 attributes correspond to 5 sets of model types: 3 sets of population growth models and 2 sets of predation models.

The assumptions from which the attributes were generated form domains of values. The resulting domains of the aforementioned attributes are summarised in table 5.

The activity constraints in the aDCSP describe the conditions that instantiate the subject of the assumptions that correspond to an attribute. Since each participant or relation has a label in the model space, a minimal set of assumptions under which it becomes part of the emerging model is available. When a participant or relation is the subject of an assumption, this label explicitly describes the sets of assumptions under which the attribute that corresponds to that subject should be activated. By translating the label of a subject into sets of attribute-value assignments, the antecedents of the activity constraints are constructed.

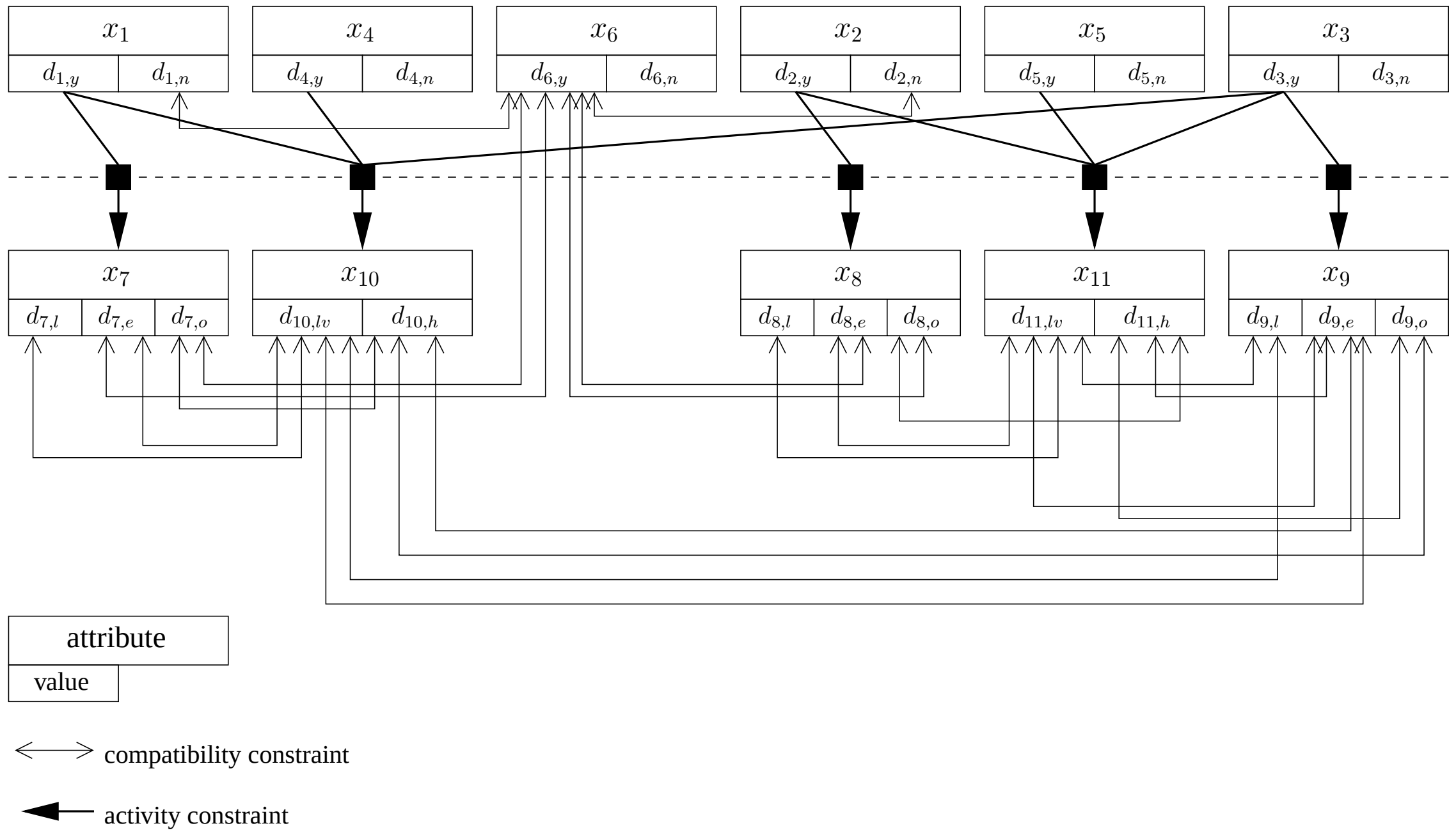

Figure 12: aDCSP derived from the models space reflecting the 1 predator and 2 competing prey scenario

In this example, the relevance assumptions (attributes $x_1, \ldots, x_6$) take their subjects from the scenario, and hence, they are always active. The attributes related to the model assumptions for population growth are active if the corresponding assumptions denoting relevance of population growth are true. That is,

$$x_1 : d_{1,y} \rightarrow \text{active}(x_7)$$
$$x_2 : d_{2,y} \rightarrow \text{active}(x_8)$$
$$x_3 : d_{3,y} \rightarrow \text{active}(x_9)$$

The attributes related to the assumptions about the predation models are active if the corresponding assumptions denoting relevance of predation, and the assumptions describing relevance of population growth, are true for the populations involved in the predation relation. That is,

$$x_1 : d_{1,y} \wedge x_3 : d_{3,y} \wedge x_4 : d_{4,y} \rightarrow \text{active}(x_{10})$$
$$x_2 : d_{2,y} \wedge x_3 : d_{3,y} \wedge x_5 : d_{5,y} \rightarrow \text{active}(x_{11})$$

Figure 12 shows a graphical representation of these activity constraints.

The compatibility constraints correspond directly to the inconsistencies in the nogood node. These inconsistencies have been discussed in the previous section and are depicted in Figure 12.

Once the aDCSP is constructed, preferences may be attached to attribute-value assignments. Suppose that preferences are only assigned to the standard population modelling choices, i.e. expo-nential growth, logistic growth, lotka-volterra predation and holling predation, and to the relevance of competition (because only one type model has been implemented for this phenomenon). For example, the following BPQs could be employed:

$$p_{\text{exponential}} < p_{\text{logistic}}$$
$$p_{\text{lotka-volterra}} < p_{\text{holling}}$$
$$p_{\text{competition}}$$

| Attribute | Preference assignments |
|---|---|
| $x_1, \ldots, x_5$ | no preference assignments |
| $x_6$ | $P(x_6 : d_{6,y}) = p_{\text{competition}}$ |
| $x_7$ | $P(x_7 : d_{7,l}) = p_{\text{logistic}}$, $P(x_7 : d_{7,e}) = p_{\text{exponential}}$ |
| $x_8$ | $P(x_8 : d_{8,l}) = p_{\text{logistic}}$, $P(x_8 : d_{8,e}) = p_{\text{exponential}}$ |
| $x_9$ | $P(x_9 : d_{9,l}) = p_{\text{logistic}}$, $P(x_9 : d_{9,e}) = p_{\text{exponential}}$ |
| $x_{10}$ | $P(x_{10} : d_{10,h} = p_{\text{holling}}$, $P(x_{10} : d_{10,lv}) = p_{\text{lotka-volterra}}$ |
| $x_{11}$ | $P(x_{11} : d_{11,h} = p_{\text{holling}}$, $P(x_{11} : d_{11,lv}) = p_{\text{lotka-volterra}}$ |

Table 6: Preference assignments for the 1 predator and 2 competing prey problem

The logistic and Holling models are preferred over the exponential and Lotka-Volterra models because the former are generally regarded as being more accurate. Note that the preferences have been ordered in such a way that those corresponding to different phenomena are not related to one another. The justification for this ordering is that, even though the models are structurally connected (there are restrictions over which models can combined with one another), models of different phenomena inherently describe behaviours that can not be compared with one another. The preference assignments for attribute value assignments are summarised in table 6.

Solving this aDPCSP is simple. First, the attributes $x_1, \ldots, x_6$ are activated. Each of these attributes is assigned $x_i : d_{i,y}$ because that assignment maximises the potential preference. Then, the attributes $x_7, \ldots, x_{11}$ are activated. Here, attributes $x_7, \ldots, x_9$ are assigned $x_i : d_{i,l}$ because the logistic growth model has the highest preference. Finally, $x_{10}$ and $x_{11}$ are assigned $x_{10} : d_{10,h}$ and $x_{11} : d_{11,h}$ because the Holling models have the highest preference and are not inconsistent with the logistic model committed earlier. The resulting solution satisfies the following set of assumptions:

```
{(relevant growth prey1),
 (relevant growth prey2),
 (relevant growth predator),
 (relevant competition prey1 prey2),
 (relevant predation predator prey1),
 (relevant predation predator prey2),
 (model size-1 logistic),
 (model size-2 logistic),
 (model size-3 logistic),
 (model predation-phen-1 holling),
 (model predation-phen-2 holling)}
```

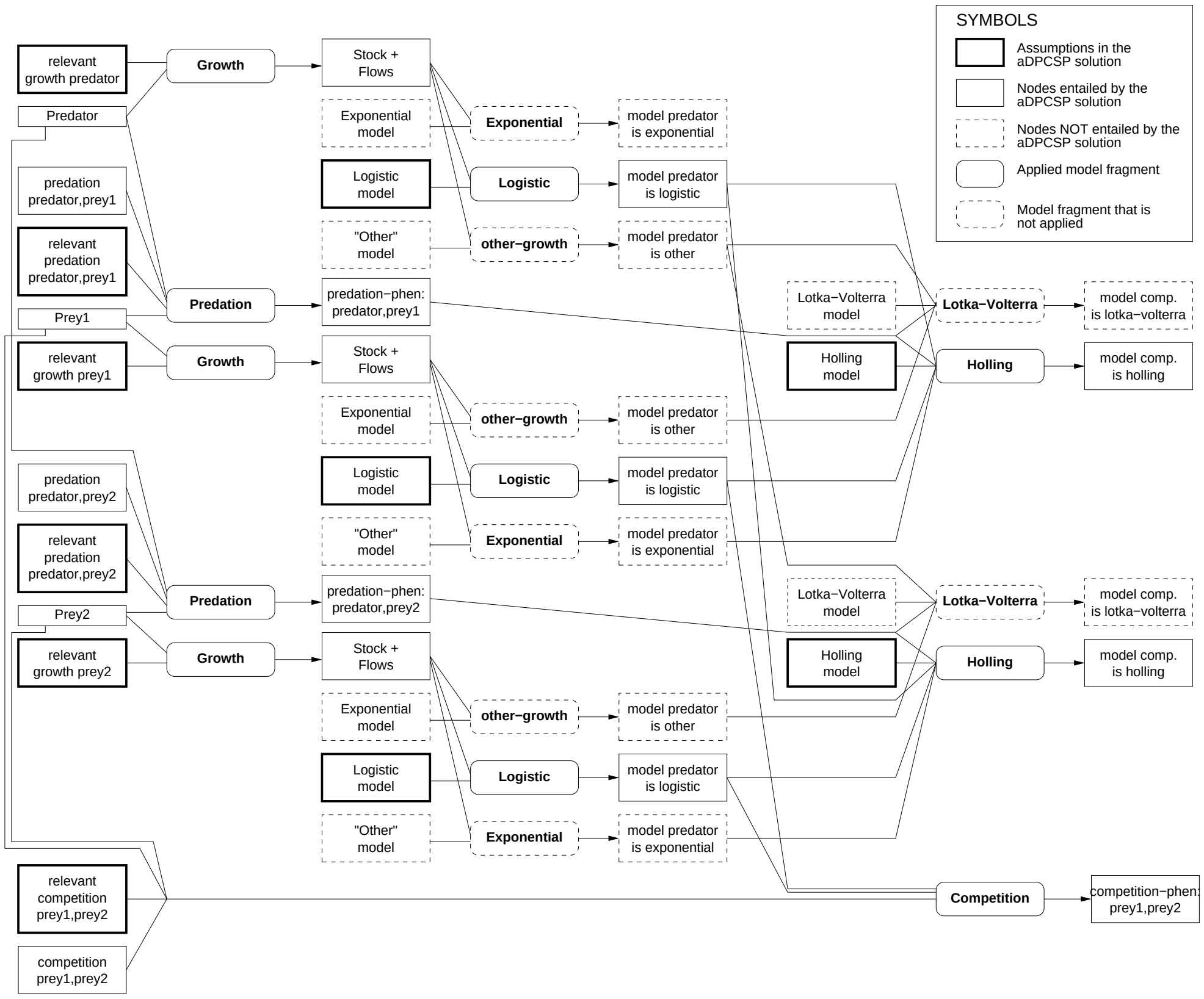

Figure 13: Deducing a scenario model from the model space, given a set of assumptions

### 4.4 Sample scenario model

Figure 13 shows how a scenario model can be deduced from the above set of assumptions by exploiting the model space. The nodes corresponding to the aforementioned assumptions and those that logically follow from the assumption set are indicated in the Figure.

When combining the participants and relations in the resulting scenario model, the model given in Figure 14 can be drawn. This model corresponds to the one that an ecologist would draw if the logistic growth and Holling predation models were regarded to be appropriate for the task at hand.

## 5. Conclusion and Future Work

This article has presented a novel approach to compositional modelling that enables the construction of models of ecological systems. This work differs from existing approaches in that it automatically translates the compositional modelling problem into an aDCSP with (order-of-magnitude) preference valuations. There are several benefits to this method.

The use of a translation algorithm that converts the compositional modelling problem into an aDCSP allows criteria to be formalised. More importantly, it also enables efficient, existing and future, aDCSP solution techniques to be effectively applied to solving compositional modelling problems.

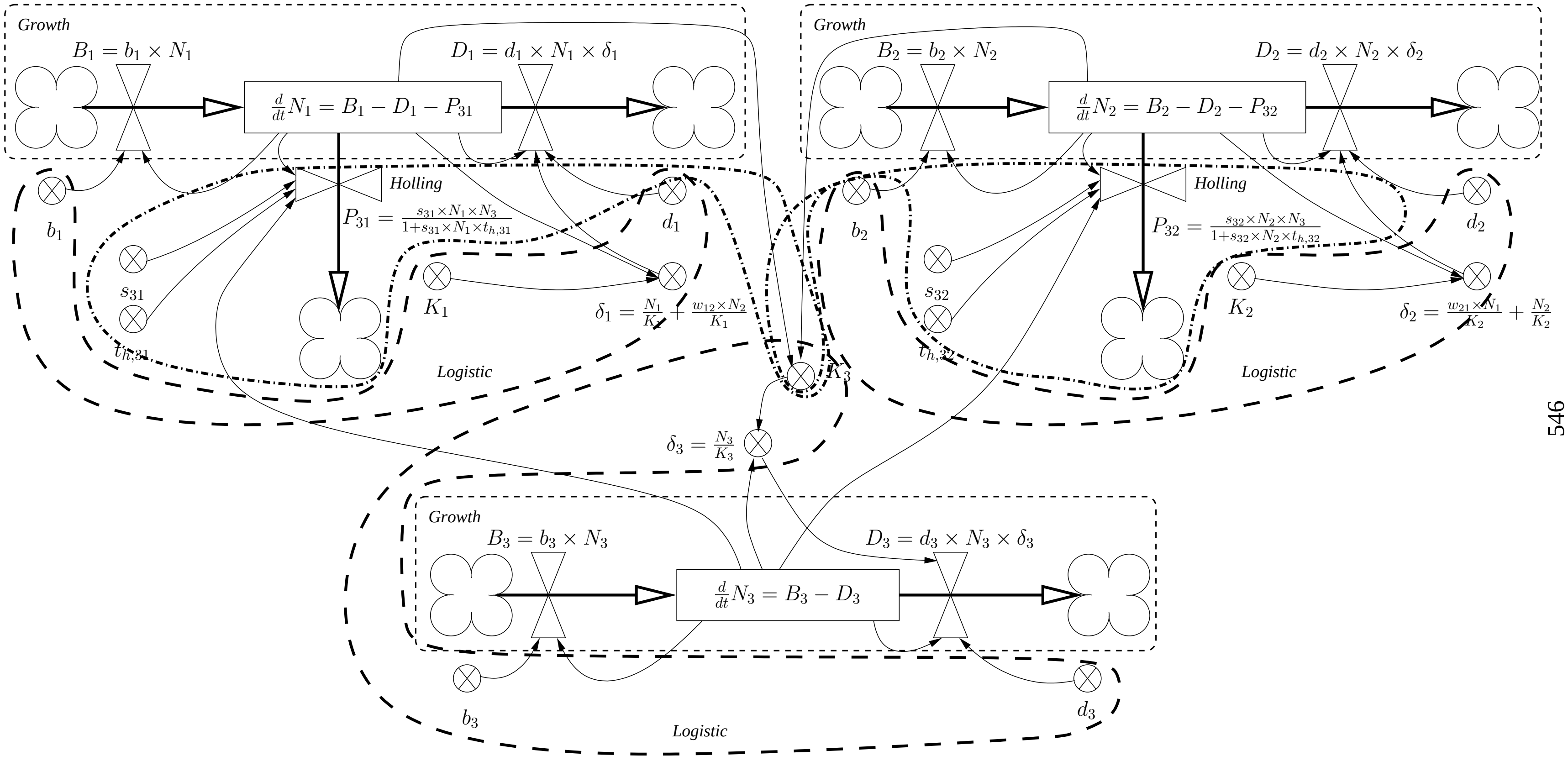

Figure 14: Sample scenario model for the 1 predator and 2 competing prey scenario

The extension of the aDCSPs with (order-of-magnitude) preferences (to form aDPCSPs) also permits the incorporation of softer requirements in the compositional modelling problem. In this paper, order-of-magnitude preferences have been employed to express the appropriateness of alternative model types for certain phenomena. While such considerations may be described by hard constraints in the physical systems domain[3], they are more subjective in less understood problem domains, such as the ecological modelling domain. The approach presented herein provides a means to capture and represent the subtlety of the flexible model design decisions.

The theoretical ideas presented in this article have been applied to real-world ecological modelling problems. In this paper, it has been demonstrated how the resultant compositional modeller can be employed to create a repository of population dynamics models. The approach has also been applied to automated model construction of large and complex ecosystems such as the MODMED model of Mediterranean vegetation (Legg et al., 1995), as reported by Keppens (2002).

There are some practical and theoretical issues that need to be addressed, however. On the practical side, the types of ecological model design decisions, as represented by the assumptions and assumption classes, and as supported by the inference mechanisms, should be extended. Ecological systems tend to involve interrelated populations of individuals, instead of functional compositions of individual components as with physical systems. One particularly important type of design decision in ecological modelling is therefore granularity. This requires the introduction of novel representation formalisms and inference mechanisms such as aggregation and disaggregation. Initial work for considering populations as single entities and for dividing such entities into sub-populations when necessary has been carried out (Keppens & Shen, 2001a). Integration of such work into the present aDPCSP framework requires further investigation.

On the theoretical side, the analysis of the complexity of the present approach is rather informal. Much remains to be done in this regard, especially when comparing to the complexity of existing compositional modellers. For this comparison, additional work will be required to adapt the current translation procedure to suit existing compositional modelling problems. Most compositional modellers are of exponential complexity, however. As they employ problem-specific solution algorithms, little is known about opportunities for improving their efficiency. This work hopes to be a first step toward further understanding this important issue.

## Acknowledgments

This work is partly supported by the UK-EPSRC grant GR/S63267. The first author has also been supported by a College of Science and Engineering scholarship at the University of Edinburgh. We are very grateful to Robert Muetzelfeldt for helpful discussions and assistance in the research reported, whilst taking the full responsibility of the views expressed here. Thanks also go to the anonymous referees for their constructive comments which are very useful in revising the earlier version of this paper.

3. These are the so-called operating conditions, stating the range of values of certain variables within which the use of certain assumptions is permitted.